\documentclass{iopjournal}

\usepackage{amsmath,amssymb,amsfonts}
\usepackage{graphicx}
\usepackage{booktabs}
\usepackage{multirow}
\usepackage{array}
\usepackage{siunitx}
\usepackage{algorithm}
\usepackage{algorithmic}
\usepackage{url}
\usepackage[authoryear,round]{natbib}
\usepackage{subfig}
\usepackage{bm}
\usepackage{mathtools}
\usepackage{tabularx}
\usepackage{float}

\hypersetup{
  hidelinks,
  pdftitle={A Multidimensional Data-Driven Hybrid Transformer Framework for Non-invasive Continuous Blood Pressure Prediction},
  pdfauthor={Yuexin Ma, Jingqi Hou, Yuxuan Kang, Zhaoying Liu}
}

\usepackage{microtype}
\usepackage{tikz}
\usetikzlibrary{shapes.geometric, arrows.meta, positioning}
\newcommand{\justifytext}{\leftskip=0pt\rightskip=0pt\spaceskip=0pt\xspaceskip=0pt\parfillskip=0pt plus 1fil\emergencystretch=2em\relax}

\makeatletter
\fancyhfoffset[L]{0pt}
\renewcommand{\headrulewidth}{0pt}
\renewcommand{\articletype}[1]{%
  \vspace*{-4mm}%
  \noindent{\scriptsize\sffamily\bfseries PREPRINT}\par
  \vspace{4mm}%
}
\makeatother

\begin{document}

\articletype{Paper}

\title{A Multidimensional Data-Driven Hybrid Transformer Framework for Non-invasive Continuous Blood Pressure Prediction}

\author{Yuexin Ma$^{1,\dagger}$, Jingqi Hou$^{1,\dagger}$, Yuxuan Kang$^{1,\dagger}$ and Zhaoying Liu$^{1,*}$}

\affil{$^1$College of Computer Science, Beijing University of Technology, Beijing, China}

\affil{$^\dagger$These authors contributed equally to this work.}

\affil{$^*$Author to whom any correspondence should be addressed.}

\email{zhaoying.liu@bjut.edu.cn}

\affil{\textbf{Other author e-mails:} Yuexin Ma: \href{mailto:andrain@emails.bjut.edu.cn}{andrain@emails.bjut.edu.cn}; Jingqi Hou: \href{mailto:lcstate@emails.bjut.edu.cn}{lcstate@emails.bjut.edu.cn}; Yuxuan Kang: \href{mailto:23070211@emails.bjut.edu.cn}{23070211@emails.bjut.edu.cn}}

\affil{\textbf{ORCID iDs:} Yuexin Ma: \href{https://orcid.org/0009-0007-9705-2374}{0009-0007-9705-2374}; Jingqi Hou: \href{https://orcid.org/0009-0002-4213-2661}{0009-0002-4213-2661}; Yuxuan Kang: \href{https://orcid.org/0009-0003-5020-1417}{0009-0003-5020-1417}; Zhaoying Liu: \href{https://orcid.org/0000-0001-6991-0123}{0000-0001-6991-0123}}


\begin{abstract}
\justifytext
\textbf{Objective.} To develop and evaluate a cuffless continuous blood pressure (BP) estimator using temporal physiological and demographic features. We propose a hybrid Transformer framework to estimate diastolic and systolic BP from ECG/PPG-derived feature sequences. \textbf{Approach.} Rather than raw waveforms, the framework models 10-step sequences of six physiological descriptors and two demographic covariates. A Multi-Source Temporal Encoder Module combines Transformer, Kolmogorov--Arnold Network, and XGBoost branches to capture complementary temporal, nonlinear, and tabular information. A Dynamic Conditional Fusion-Decoder applies differential multi-head attention, token-weighted aggregation, and gated residual correction. A robust composite objective jointly optimizes DBP and SBP. \textbf{Main results.} Using the MIMIC-III Waveform and Clinical Databases, the source pool comprised 28,486 waveform segments from 203 subjects, and feature generation retained 53,621 observations from 166 subjects. On 2,431 segment-level held-out test windows, mean error~$\pm$~standard deviation was \SI{0.41}{mmHg}~$\pm$~\SI{3.74}{mmHg} for diastolic BP and \SI{-1.60}{mmHg}~$\pm$~\SI{5.95}{mmHg} for systolic BP, with 95\% limits of agreement of [-6.93, 7.74] and [-13.25, 10.06] mmHg, respectively. The proportions within \SI{10}{mmHg} were 98.48\% and 94.36\%. The framework achieved the lowest standard deviations and narrowest limits of agreement among the locally retrained baselines. \textbf{Significance.} The feature-sequence fusion framework improved agreement with reference BP and fell within numerical AAMI and BHS Grade~A thresholds on this split. This retrospective analysis is not formal device validation; subject-disjoint and external evaluation remain necessary before clinical use.
\end{abstract}

\justifytext

\section{Introduction}
Hypertension is a major modifiable risk factor for cardiovascular disease and stroke worldwide \citep{mcevoy2024esc, zhou2021global}. Conventional non-invasive blood pressure (BP) measurement mainly relies on cuff-based techniques, including auscultatory and oscillometric methods \citep{muntner2019bp, meidert2018noninvasive}. However, these methods provide only intermittent measurements and are limited in long-term continuous physiological monitoring owing to cuff-related discomfort, device bulkiness, and operational constraints \citep{meidert2018noninvasive, pilz2024cuff, hua2024wearable, min2025wearable}. To address these limitations, non\mbox{-}invasive cuffless monitoring technologies have garnered significant attention because of their potential to revolutionize traditional practices by offering undisturbed, sustainable measurements \citep{hua2024wearable,min2025wearable,zhao2023emerging}. In particular, the pulse transit time (PTT)---defined as the temporal delay between the cardiac electrical activation (ECG R\mbox{-}peak) and the peripheral pulse arrival (PPG foot)---has been widely validated as a surrogate marker negatively correlated with BP \citep{mukkamala2015ptt,block2020cptt,ding2016pir}.

Nonetheless, BP estimation relying solely on PTT is susceptible to inter\mbox{-}subject variability and environmental interferences, often requiring frequent calibration to compensate for these offsets \citep{mukkamala2015ptt,zhao2023emerging}. To enhance precision, recent studies have introduced auxiliary physiological descriptors; for instance, \citet{ding2016pir} proposed the PPG intensity ratio (PIR) to complement PTT, significantly improving estimation accuracy and dynamic tracking capabilities. Concurrently, machine learning and deep learning paradigms have been deployed to extract features directly from signal waveforms. While some approaches utilize regression algorithms on statistical features to meet Association for the Advancement of Medical Instrumentation (AAMI) standards \citep{kachuee2017aami,huang2023cnnsvr}, others employ deep models like Convolutional Neural Networks (CNNs) and Recurrent Neural Networks (RNNs) to learn representations from raw ECG and PPG signals \citep{slapnicar2019ppg,tang2024kd}. However, despite their initial success, these existing deep models often struggle with generalization across diverse populations, lack robustness against motion artifacts, and suffer from poor interpretability \citep{zhao2023emerging,elgendi2023icu,mehta2024challenges}.

To overcome these challenges, we propose a multidimensional data\mbox{-}driven pipeline that fuses ECG/PPG-derived physiological descriptors with an improved hybrid Transformer architecture. Rather than feeding raw waveform samples directly to the network, the method models consecutive records of timing, cardiac, PPG morphology, and demographic features. The source ECG, PPG, and continuous arterial blood pressure (ABP) signals were obtained from the MIMIC\mbox{-}III Waveform Database, and age and body weight were linked from the MIMIC\mbox{-}III Clinical Database v1.4 \citep{moody2020mimic3wdb,johnson2016mimic,johnson2016mimicdb}.

The main contributions of this work are summarized as follows.

First, we propose a Multi\mbox{-}Source Temporal Encoder Module (MSEM) to process multivariate physiological feature sequences. By integrating a Transformer encoder with parallel Kolmogorov--Arnold Network (KAN) and XGBoost (XGB) branches, the MSEM captures complementary temporal, non\mbox{-}linear, and tabular representations of hemodynamic variation \citep{liu2024kan,chen2016xgboost}.

Second, we introduce a Dynamic Conditional Fusion\mbox{-}Decoder (DCFD) to unify multi\mbox{-}branch representations. It employs a differential multi\mbox{-}head attention mechanism to reduce feature redundancy and enhance cross\mbox{-}branch fusion efficiency \citep{difftransformer2024}.

Third, we formulate a composite robust point-regression objective as a weighted sum of complementary terms rather than treating them as separate training stages. Learnable target-scale parameters balance the DBP and SBP squared-error terms; scheduled focal-style and quantile penalties emphasize large and asymmetric residuals; an early-epoch Huber term stabilizes initial optimization; and KAN, target-scale, and residual-correction terms provide regularization and auxiliary supervision \citep{kendall2018uncertainty,lin2017focal,koenker1978quantiles,huber1964}. Evaluated on the MIMIC\mbox{-}III dataset \citep{moody2020mimic3wdb}, this approach falls within the numerical AAMI mean-error and SD criteria and the numerical BHS Grade~A percentage thresholds on the segment-level held-out test set \citep{iso8106022018,iso8106032022,obrien1993bhs}.

\section{Related Work}

Cuffless BP estimation traditionally exploits the inverse correlation between pulse transit time (PTT) and BP \citep{mukkamala2015ptt}. While PTT remains a fundamental physiological marker, PTT\mbox{-}only models are sensitive to individual differences and vascular compliance changes \citep{mukkamala2015ptt,zhao2023emerging}. To enhance precision, recent literature augments PTT with morphological and intensity descriptors, such as the pulse intensity ratio (PIR) \citep{ding2016pir}, and applies deep learning paradigms to extract representations directly from ECG and PPG waveforms \citep{huang2023cnnsvr,tang2024kd,zhao2023emerging}. Existing studies suggest that robust preprocessing, multimodal feature fusion, and attention\mbox{-}based modeling are promising directions for improving cuffless BP estimation performance.

\subsection{Evolution of Deep Learning Architectures}
The progression of BP prediction models has shifted from feature engineering to end\mbox{-}to\mbox{-}end learning, focusing on capturing \textit{spatial morphological features} and \textit{temporal dependencies}.

\textbf{Convolutional Neural Networks (CNNs).} Convolutional Neural Networks serve as powerful extractors for local waveform patterns. Variants like 1D\mbox{-}CNN and ResNet have proven effective in learning abstract features such as systolic upstroke time and notch width \citep{slapnicar2019ppg,liu2025resnetbilstm}. Notably, architectures like DSRUnet introduce sparse residual connections and deep supervision to prevent gradient vanishing, achieving AAMI\mbox{-}compliant results on public datasets \citep{lai2024dsrunet}. U-Net-based waveform reconstruction and lightweight Squeeze U-Net variants have also been explored for PPG-based BP estimation \citep{athaya2021unet,athaya2022squeezeunet}. A recent image-encoding alternative, GHC-Net, transforms PPG signals into Gramian angular fields and combines hybrid dilated convolutions with channel attention; its reported task is BP-category classification rather than continuous-value regression \citep{cheng2026ghcnet}. Parallel or multi\mbox{-}stream CNN-based frameworks have also been employed to process signal derivatives (e.g., velocity and acceleration PPG), thereby mimicking clinically relevant waveform interpretation \citep{slapnicar2019ppg}.

\textbf{Recurrent Neural Networks (RNNs).} To model the dynamic nature of BP, Recurrent Neural Networks, particularly LSTMs and GRUs, are widely used to capture historical dependencies and autonomic regulation \citep{zhao2023emerging}. For instance, hybrid models combining CNNs for spatial extraction and Bi\mbox{-}LSTMs/Conv\mbox{-}LSTMs for temporal modeling have demonstrated high accuracy \citep{jeong2021cnnlstm,kamanditya2024convlstm,liu2025resnetbilstm,tanveer2018annlstm}. In 2026, LAST-CBPM combined wavelet-based PPG denoising, temporal self-attention, BiLSTM modeling, and subject-specific residual calibration for quasi-continuous BP monitoring on MIMIC-IV and an independent in-house dataset \citep{song2026last}. However, RNNs entail serial computation and may be less effective at modeling very long-range dependencies, which can limit their ability to capture ultra\mbox{-}long temporal patterns such as circadian rhythms \citep{vaswani2017attention}.

\textbf{Transformers and Generative Models.} Recently, Transformer architectures utilizing self\mbox{-}attention have emerged as strong performers in cuffless BP estimation and waveform synthesis \citep{tian2025pctn,zheng2025utransbpnet,nawaz2024transformerabp}. Models like UTransBPNet combine U\mbox{-}Net backbones with cross\mbox{-}attention mechanisms to track hemodynamic changes in dynamic scenarios \citep{zheng2025utransbpnet}. Recent 2026 studies further combine attention with complementary signal representations: MCAFNet uses ConvNeXt and Transformer branches with multi\mbox{-}scale cross\mbox{-}attention to reconstruct continuous ABP from PPG, whereas FFDM-GAT-Transformer couples Fourier-decomposition-derived PPG features with graph attention and a Transformer encoder for point BP estimation \citep{tian2026mcaf,batra2026ffdm}. Generative Adversarial Networks (GANs) have also been explored for ``waveform translation'' (mapping PPG to ABP), as exemplified by PPG2ABP and CycleGAN\mbox{-}based reconstruction approaches \citep{ibtehaz2022ppg2abp,mehrabadi2022cyclegan}. Diffusion models, such as TDSTF, have likewise been investigated for blood pressure time\mbox{-}series forecasting and probabilistic generation in ICU settings, helping address data sparsity and uncertainty \citep{chang2024tdstf}.
In another 2026 multimodal design, \citet{ma2026gaf} encoded ECG and PPG signals using Gramian angular fields and combined convolutional, recurrent, and temporal\mbox{-}attention modules to capture complementary spatial\mbox{-}temporal patterns for continuous BP prediction.

\subsection{Multimodal Data Fusion Strategies}
Single\mbox{-}modality approaches are vulnerable to motion artifacts and physiological noise. Consequently, effective fusion of ECG, PPG, and other complementary modalities (including bioimpedance) is increasingly regarded as a key strategy for improving robustness against artifact\mbox{-}induced degradation and sensor\mbox{-}specific noise \citep{zhao2023emerging,min2025wearable}.

\textbf{Feature\mbox{-}, Decision\mbox{-}, and Model\mbox{-}level Fusion.} Feature\mbox{-}level fusion typically involves concatenating handcrafted descriptors (PTT, PIR) or deep embeddings prior to regression. While straightforward, this requires strict temporal synchronization and may overlook complex nonlinear cross\mbox{-}modal interactions \citep{stahlschmidt2022fusion,duan2024fusion,ma2023multiparameter}. Decision\mbox{-}level fusion aggregates outputs from independent classifiers (e.g., SVM, RF) via voting or stacking \citep{stahlschmidt2022fusion,duan2024fusion}. In contrast, model\mbox{-}level fusion jointly learns cross\mbox{-}modal interactions within a shared architecture, often using Mixture\mbox{-}of\mbox{-}Experts (MoE) or attention mechanisms. For example, recent wearable systems use gating or attention networks to dynamically recalibrate sensor contributions according to contextual cues such as temperature, contact force, or motion state, thereby improving stability across resting and exercise conditions \citep{li2025fusensering,chen2025sixmodal}.

\subsection{Gap Analysis: Generalization and Calibration}
Despite architectural advances, three key gaps remain between academic benchmarks and real\mbox{-}world deployment \citep{zhao2023emerging,min2025wearable}.
A recent systematic review further identified protocol realism, calibration drift, population diversity, and fairness\mbox{-}aware reporting as major barriers to trustworthy real\mbox{-}world cuffless BP monitoring \citep{cisnal2026trustworthy}.

\textbf{Subject Variability.} The mapping from PPG to BP is highly subject\mbox{-}specific, governed by arterial stiffness, vascular compliance, and age. Most deep models can degrade markedly in subject\mbox{-}independent or out\mbox{-}of\mbox{-}distribution evaluation \citep{zhao2023emerging,moulaeifard2025generalizable,mehta2024challenges}. While domain adaptation techniques exist, they often introduce additional methodological and computational complexity \citep{moulaeifard2025generalizable}.
Recent work has attempted to mitigate this variability through personalized adaptation. \citet{tang2026fewshot} combined self\mbox{-}supervised representation learning with few\mbox{-}shot subject\mbox{-}specific adaptation using only a small number of labeled samples. Although promising, the requirement for subject\mbox{-}specific labeled BP data remains a practical obstacle to fully calibration\mbox{-}free deployment.

\textbf{Calibration Dependency.} Many recent solutions still require periodic calibration using cuff devices to determine baseline offsets \citep{mukkamala2015ptt,kasbekar2023optimizing,min2025wearable}. Calibration\mbox{-}free models incorporating demographic data (age, BMI) have been explored, but robustness across diverse populations remains an open issue, particularly given continued concerns about skin\mbox{-}tone\mbox{-}related limitations in PPG sensing and validation \citep{liang2026cspm,colvonen2021response}.

\textbf{Computational Efficiency.} High\mbox{-}performing attention\mbox{-}based models can impose non\mbox{-}trivial memory, latency, and power costs, posing challenges for deployment on resource\mbox{-}constrained wearable edge devices \citep{min2025wearable,pankaj2025edge}.

Addressing these limitations, our proposed method introduces a hybrid framework. We combine Kolmogorov--Arnold Networks (KAN) with a differential-attention-based fusion mechanism to enhance multi\mbox{-}branch feature integration \citep{liu2024kan,difftransformer2024}.

\section{Methodology}

\subsection{Problem Definition}
Let $\mathcal{D} = \{(\mathbf{X}_i, \mathbf{y}_i)\}_{i=1}^{N}$ denote the model-ready dataset with $N$ prediction samples. Each input $\mathbf{X}_i \in \mathbb{R}^{L \times F}$ contains $L=10$ consecutive feature records and $F=8$ variables: pulse transit time (PTT), heart rate, PPG foot-to-peak interval, maximum PPG slope, reflection index (RI), systolic-to-diastolic timing (SD time), body weight, and age. The target $\mathbf{y}_i = [y_{dbp}, y_{sbp}]^\top \in \mathbb{R}^2$ contains the DBP and SBP reference values of the subsequent record. Our goal is to learn a mapping $\mathcal{F}: \mathbf{X}_i \to \hat{\mathbf{y}}_i$ that minimizes prediction error against the ABP-derived references.

\subsection{Overall Framework}
\begin{figure*}[t!] 
    \centering
    \includegraphics[width=\linewidth]{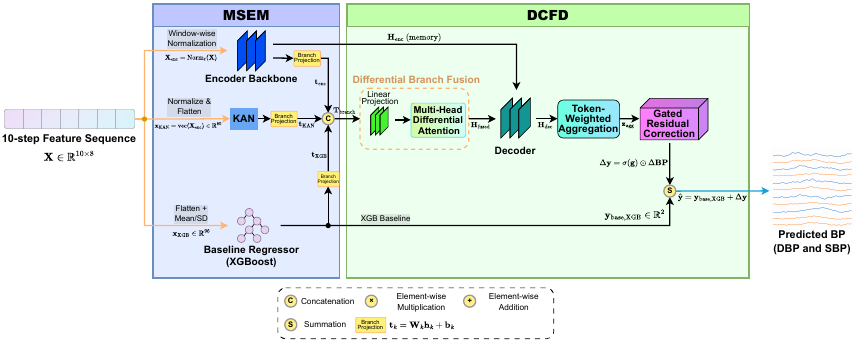} 
    
    \caption{\textbf{Architecture of the proposed hybrid framework for continuous blood pressure prediction.}
    Given a standardized feature window $\mathbf{X}\in\mathbb{R}^{10\times8}$, the \textbf{Multi-Source Temporal Encoder Module (MSEM)} processes complementary temporal, nonlinear, and tabular views using Transformer, KAN, and XGBoost branches, respectively. The encoder states $\mathbf{H}_{\mathrm{enc}}$ serve as decoder memory, while branch-specific projections $\mathbf{t}_k=\mathbf{W}_k\mathbf{h}_k+\mathbf{b}_k$ map the three representations to the common tokens $\mathbf{t}_{\mathrm{enc}}$, $\mathbf{t}_{\mathrm{KAN}}$, and $\mathbf{t}_{\mathrm{XGB}}$. Their concatenation $\mathbf{T}_{\mathrm{branch}}$ is processed by the \textbf{Dynamic Conditional Fusion-Decoder (DCFD)}, producing the fused representation $\mathbf{H}_{\mathrm{fused}}$, decoder output $\mathbf{H}_{\mathrm{dec}}$, token-weighted aggregate $\mathbf{z}_{\mathrm{agg}}$, and neural correction $\Delta\mathbf{y}$. In parallel, XGBoost supplies the two-dimensional baseline $\mathbf{y}_{\mathrm{base,XGB}}$; the final prediction is $\hat{\mathbf{y}}=\mathbf{y}_{\mathrm{base,XGB}}+\Delta\mathbf{y}$. The exact implemented decomposition of $\Delta\mathbf{y}$ is given in equation~\ref{eq:prediction}.}
    \label{fig:framework}
\end{figure*}

\begin{figure}[htbp]
    \centering
    \begin{tikzpicture}[
        node distance=0.8cm,
        auto,
        block/.style={
            rectangle,
            draw,
            text width=20em,
            text centered,
            rounded corners,
            minimum height=3em,
            line width=0.8pt
        },
        line/.style={
            draw,
            -Latex,
            thick
        }
    ]

    \node [block] (preprocess) {
        \textbf{Upstream Signal Preparation}\\
        \small Filtering and robust normalization;\\
        polarity correction and fiducial detection
    };

    \node [block, below=of preprocess] (features) {
        \textbf{Model-Ready Feature Sequences}\\
        \small Six physiological descriptors, age, and weight; \\
        segment split, training-only scaling, and $10\times8$ windows
    };

    \node [block, below=of features] (model) {
        \textbf{Hybrid Transformer Modeling}\\
        \small MSEM and DCFD with the composite objective $\mathcal{L}_{total}$
    };

    \node [block, below=of model] (train) {
        \textbf{Model Optimization \& Evaluation}\\
        \small RAdam--Lookahead with cosine annealing; \\
        segment-level validation and testing
    };

    \path [line] (preprocess) -- (features);
    \path [line] (features) -- (model);
    \path [line] (model) -- (train);

    \end{tikzpicture}

    \caption{\textbf{The overall workflow of the proposed continuous blood pressure prediction system.} Source waveform processing provides the physiological provenance of the eight retained variables. The prediction stage operates on standardized 10-step feature sequences rather than raw ECG/PPG waveform samples.}
    \label{fig:workflow}
\end{figure}
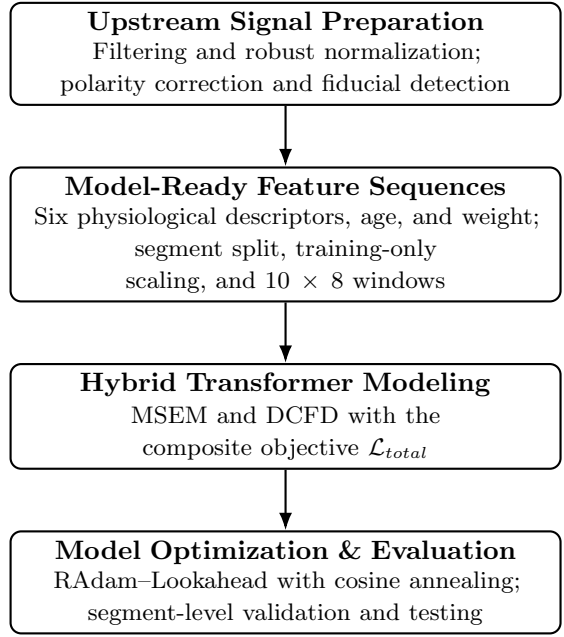

The proposed framework is a multidimensional data-driven pipeline designed for continuous cuffless BP monitoring. 
As illustrated in Fig.~\ref{fig:framework} and Fig.~\ref{fig:workflow}, the system integrates four tightly coupled components: 
(1) source waveform and metadata construction, 
(2) upstream derivation of physiological descriptors, 
(3) model-ready feature-sequence construction, and 
(4) hybrid Transformer-based prediction with a composite optimization objective. 
In particular, the model consists of a Multi-Source Temporal Encoder Module (MSEM) and a Dynamic Conditional Fusion-Decoder (DCFD), which together capture temporal dependencies, nonlinear hemodynamic characteristics, and effective cross-branch feature interactions.

For clarity, Fig.~\ref{fig:workflow} summarizes the complete execution flow from upstream signal preparation to model evaluation.

\subsection{Hybrid Transformer Architecture}
The deep learning backbone consists of two modules: the Multi-Source Temporal Encoder Module (MSEM) and the Dynamic Conditional Fusion-Decoder (DCFD).

\subsubsection{Multi-Source Temporal Encoder Module (MSEM)}
The encoding stage is designed to capture both sequential dependencies and statistical distributions. Before window construction, missing observations are removed and the eight input variables are standardized using statistics fitted on the training partition. Each batch therefore contains $\mathbf{X} \in \mathbb{R}^{B \times 10 \times 8}$. Within the model, each window is additionally normalized along the temporal dimension for the neural branches. The KAN input is the resulting 80-dimensional flattened window. The XGBoost input is formed from the globally standardized window flattened to 80 dimensions and concatenated with eight feature-wise temporal means and eight temporal standard deviations, yielding 96 dimensions.

The backbone utilizes a Transformer encoder to extract temporal hidden states $\mathbf{H}$. 
To leverage the strengths of different modeling paradigms, a branch generator produces three parallel token representations:
\begin{itemize}
    \item \textbf{Main Head:} a linear projection of the mean-pooled Transformer representation, providing a temporal feature-sequence estimate;
    \item \textbf{KAN Head:} a Kolmogorov--Arnold Network operating on the normalized 80-dimensional flattened window to model explicit nonlinear physiological relationships;
    \item \textbf{XGB Head:} an XGBoost representation operating on the 96-dimensional flattened-window statistics to provide a robust tabular baseline estimate.
\end{itemize}

Unlike conventional Multi-Layer Perceptrons (MLPs) with fixed node-wise activation functions, Kolmogorov--Arnold Networks (KANs) place learnable activation functions on edges. 
For an input vector $\mathbf{x} \in \mathbb{R}^{d_{in}}$, a KAN layer is defined as a matrix of learnable one-dimensional functions $\Phi = \{\phi_{q,p}\}$, where $\phi_{q,p}$ connects the $p$-th input to the $q$-th output node. 
The computation is expressed as:
\begin{equation}
    x_q^{l+1} = \sum_{p=1}^{d_{in}} \phi_{q,p}(x_p^l),
    \qquad q = 1, \dots, d_{out}
\end{equation}

where $\phi_{q,p}(\cdot)$ is typically parameterized as a B-spline or a weighted sum of basis functions. 
This design enables the network to better approximate the highly nonlinear hemodynamic variations involved in BP prediction.

\subsubsection{Dynamic Conditional Fusion-Decoder (DCFD)}
The fusion-decoder stage is designed to integrate heterogeneous representations produced by different modeling branches. 
Given the branch tokens $\mathbf{t}_i$, we employ a Multi-Head Differential Attention (MHDA) module to enhance informative interactions while suppressing redundant responses across branches.

Unlike standard self-attention, MHDA computes two distinct similarity maps $(A_1, A_2)$ and subtracts them, i.e.,
\begin{equation}
    A = A_1 - \lambda A_2,
\end{equation}
so as to suppress redundant common-mode information while emphasizing salient variations.

The attended features are then aggregated into a fused representation $\mathbf{H}_{fused}$ for final prediction. 
To ensure training stability and gradual integration of complex features, we employ a gated residual fusion strategy:
\begin{equation}
    \label{eq:prediction}
    \hat{\mathbf{y}} = \mathbf{y}_{\mathrm{base,XGB}}
    + \sigma(\mathbf{g}) \odot r(s)\,\mathrm{Linear}(\mathbf{H}_{\mathrm{fused}})
    + \sum_k \gamma_k \mathbf{t}_k^{\mathrm{res}}
\end{equation}
where $\mathbf{y}_{\mathrm{base,XGB}}$ is the robust two-dimensional XGBoost baseline estimate, $\sigma(\mathbf{g})$ is the sigmoid gate, $r(s)$ is the residual ramp coefficient at training step $s$, and $\gamma_k$ weights the residual token $\mathbf{t}_k^{\mathrm{res}}$ from branch $k$. The terms following $\mathbf{y}_{\mathrm{base,XGB}}$ jointly constitute the aggregate neural correction $\Delta\mathbf{y}$ shown compactly in Fig.~\ref{fig:framework}; the neural branches therefore mainly learn residual corrections to the statistical baseline.

\subsection{Dataset and Data Preparation}
\label{subsec:dataset_preparation}
We used synchronized ECG, PPG, and ABP segments from the MIMIC-III Waveform Database v1.0 \citep{moody2020mimic3wdb,johnson2016mimic}, resampled to \SI{125}{Hz}. Age and weight were linked from the MIMIC-III Clinical Database v1.4 \citep{johnson2016mimicdb}. Table~\ref{tab:cohort_summary} summarizes cohort composition before and after signal-quality screening and feature generation.

\begin{table}[htbp]
  \centering
  \caption{Composition of the source metadata pool and final analytical cohort. Percentages are calculated within each row.}
  \label{tab:cohort_summary}
  \small
  \setlength{\tabcolsep}{3pt}
  \renewcommand{\arraystretch}{1.08}
  \begin{tabularx}{\columnwidth}{@{}p{0.18\columnwidth}p{0.20\columnwidth}>{\centering\arraybackslash}p{0.11\columnwidth}>{\centering\arraybackslash}p{0.19\columnwidth}>{\centering\arraybackslash}X@{}}
    \toprule
    \textbf{Stage} & \textbf{Statistical unit} & \textbf{Total, $n$} & \textbf{Male, $n$ (\%)} & \textbf{Female, $n$ (\%)} \\
    \midrule
    \multirow{2}{*}{Source pool}
      & Subjects & 203 & 115 (56.65) & 88 (43.35) \\
      & Waveform segments & 28,486 & 16,791 (58.94) & 11,695 (41.06) \\
    \midrule
    \multirow{2}{*}{Analytical cohort}
      & Subjects & 166 & 104 (62.65) & 62 (37.35) \\
      & Observations & 53,621 & 43,620 (81.35) & 10,001 (18.65) \\
    \bottomrule
  \end{tabularx}
\end{table}

Because the exported model-ready table omitted subject identifiers and sex labels, provenance was reconstructed from the source metadata and waveform archive. Unique age--weight combinations resolved 52,890 observations; the remaining 731 observations from three shared combinations were assigned by matching continuous SBP--DBP pairs to raw MAT-file labels at $10^{-9}$ precision. This mapping was used only for cohort accounting, not as a model input or partitioning variable.

Feature generation required finite BP references and detectable ECG/PPG fiducials. Complete-case filtering then removed 9 of 53,630 rows with missing weight. No pre-specified physiological-range, $\mathrm{SBP}>\mathrm{DBP}$, or RI-clipping rule was applied; therefore, the analytical table retained two rows with $\mathrm{SBP}\leq\mathrm{DBP}$, approximate minima of \SI{3.8}{mmHg} for DBP and \SI{50.0}{mmHg} for SBP, and extreme RI values. No retrospective exclusion was performed to preserve the original modeling pipeline. A post hoc audit of the 2,431 held-out test references found DBP and SBP ranges of 35.62--98.23 and 85.36--185.26 mmHg, respectively, with no test row satisfying $\mathrm{SBP}\leq\mathrm{DBP}$.

\subsection{Sex and Gender Considerations}
In accordance with the Sex and Gender Equity in Research (SAGER) guidelines \citep{heidari2016sager}, sex was considered as a biological variable when describing the source pool and final analytical cohort. The binary M/F field available in the source metadata was not retained as a model predictor, a partitioning variable, or a stratification factor when the model-ready table and experimental splits were created; the subsequent subject mapping was used only for cohort accounting. The higher proportion of male subjects in the final cohort reflects the composition of the available secondary ICU data after signal-quality screening and was not caused by sex-specific recruitment or exclusion criteria. Gender identity was not available in the source databases; therefore, this article reports sex rather than gender and makes no gender-based inferences. Model performance was not evaluated separately by sex, and the reported aggregate results should not be interpreted as evidence of equivalent performance across sex groups.

\subsection{Preprocessing and Signal Alignment}
\label{subsec:preprocessing}

\begin{figure}[H]
    \centering
    \includegraphics[width=0.84\linewidth]{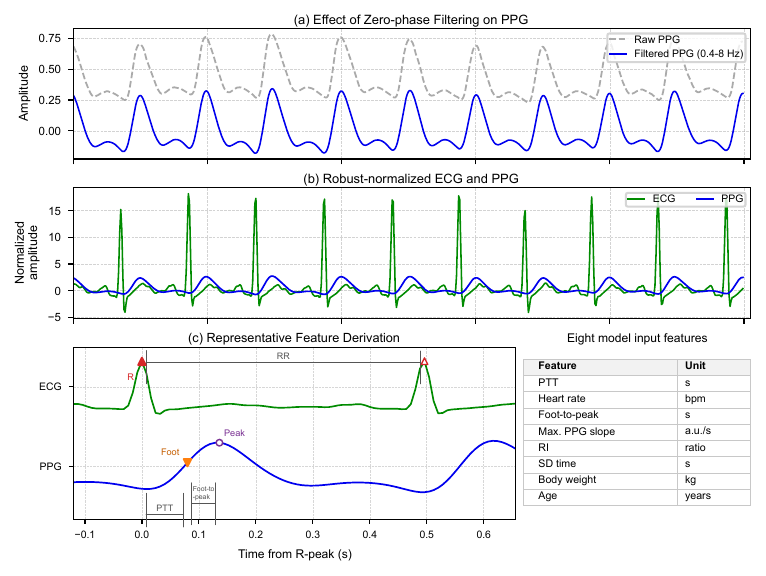}
    \caption{\textbf{Upstream signal preparation and feature provenance.}
    (a) Raw and zero-phase band-pass-filtered PPG.
    (b) Robust-normalized ECG and PPG after polarity correction.
    (c) Fiducial landmarks used to derive PTT, the foot-to-peak interval, and heart rate, together with the eight retained predictors. The waveforms illustrate feature derivation; the prediction model receives 10-step feature sequences rather than raw signals.}
    \label{fig:preprocessing}
\end{figure}

The modeling stage begins from an exported feature table and does not load raw ECG or PPG waveform arrays. The waveform processing described here documents the physiological provenance of the retained descriptors rather than the tensor representation supplied to the network. The source signals were sampled at \SI{125}{Hz}; representative upstream processing used third-order zero-phase Butterworth filtering (ECG: \SIrange{0.5}{40}{Hz}; PPG: \SIrange{0.4}{8}{Hz}), robust median/IQR normalization, and PPG polarity correction before fiducial detection. Fig.~\ref{fig:preprocessing} illustrates these operations and the timing landmarks from which the model-ready variables were derived.

ECG R-peaks and the corresponding PPG pulse feet and systolic peaks provide the timing landmarks for three retained variables. Let $t_R^{(j)}$ and $t_R^{(j+1)}$ denote the sample indices of two consecutive ECG R-peaks, and let $t_{\mathrm{foot}}$ and $t_{\mathrm{peak}}$ denote the PPG foot and systolic-peak indices associated with the $j$th cardiac cycle. With sampling frequency $f_s=\SI{125}{Hz}$, pulse transit time (PTT), the PPG foot-to-peak interval ($T_{\mathrm{foot\mbox{-}peak}}$), the RR interval ($T_{RR}$), and heart rate (HR) are defined as
\begin{equation}
\begin{aligned}
    \mathrm{PTT} &= \frac{t_{\mathrm{foot}} - t_R^{(j)}}{f_s},
    &
    T_{\mathrm{foot\mbox{-}peak}} &= \frac{t_{\mathrm{peak}} - t_{\mathrm{foot}}}{f_s},\\[3pt]
    T_{RR} &= \frac{t_R^{(j+1)} - t_R^{(j)}}{f_s},
    &
    \mathrm{HR} &= \frac{60}{T_{RR}}.
\end{aligned}
\end{equation}
Here, PTT, $T_{\mathrm{foot\mbox{-}peak}}$, and $T_{RR}$ are expressed in seconds, and HR is expressed in beats per minute.
The remaining waveform-derived inputs are the maximum PPG slope, the reflection index (RI; reflected-wave amplitude relative to the primary systolic amplitude), and SD time (the systolic-to-diastolic PPG timing interval). Age and body weight are appended as static demographic covariates. These definitions describe the eight columns retained in the final model-ready table; waveform samples and other candidate morphology descriptors are not supplied to the prediction network.

\subsection{Physiological Feature Engineering}
\label{subsec:feature_engineering}

The final analytical table contains exactly eight predictors per record. The six physiological descriptors derived upstream from synchronized ECG/PPG recordings are pulse transit time (PTT), PPG foot-to-systolic-peak interval, heart rate, maximum PPG slope, reflection index (RI), and systolic-to-diastolic timing (SD time). The two static demographic covariates are body weight and age. Table~\ref{tab:features} lists their stored variable names and definitions. No raw waveform samples, resampled waveform templates, wavelet coefficients, spectral descriptors, shapelets, AUC features, or P2 descriptors are included in the model input.

\begin{table}[htbp]
  \centering
  \caption{Predictor variables retained in the model-ready feature table.}
  \label{tab:features}
  \small
  \setlength{\tabcolsep}{4pt}
  \renewcommand{\arraystretch}{1.08}
  \begin{tabularx}{\columnwidth}{p{0.20\columnwidth} p{0.30\columnwidth} X}
    \toprule
    \textbf{Category} & \textbf{Stored variable} & \textbf{Description} \\
    \midrule
    Timing
      & \textit{ptt}
      & ECG R-peak to PPG foot interval (s) \\
    \midrule
    Timing
      & \textit{ptt\_foot\_pk}
      & PPG foot-to-systolic-peak interval (s) \\
    \midrule
    Cardiac
      & \textit{hr}
      & Heart rate (beats/min) \\
    \midrule
    PPG morphology
      & \textit{ppg\_slope\_max}
      & Normalized maximum PPG slope (a.u./s) \\
    \midrule
    PPG morphology
      & \textit{ri}
      & Reflection index (ratio) \\
    \midrule
    PPG timing
      & \textit{sd\_time}
      & Systolic-to-diastolic timing descriptor (s) \\
    \midrule
    Demographic
      & \textit{weight}
      & Body weight (kg) \\
    \midrule
    Demographic
      & \textit{age}
      & Age (years) \\
    \bottomrule
  \end{tabularx}
\end{table}

The 53,621 complete observations described above were used without additional post hoc outlier exclusion. Segment boundaries were inferred from resets or discontinuities in the exported index, and complete segments were assigned to the training, validation, and test partitions before scaling or window construction. The input scaler and output scaler were fitted only on training-segment rows. Within each segment, a unit-stride window uses 10 consecutive standardized feature records to predict the DBP and SBP values of the next record. Thus, each neural input has shape $10\times8$, while the prediction length is one record.

\subsection{Composite Robust Loss Function}
BP prediction is formulated as a two-target point-regression task requiring accurate and robust DBP and SBP estimates. The implemented network returns one DBP and one SBP estimate for each input window. We define the overall training objective as
\begin{equation}
\mathcal{L}_{total}
= \sum_{k\in\{\mathrm{DBP},\mathrm{SBP}\}}w_k\mathcal{L}_{A,k}
+ \alpha_F \mathcal{L}_F
+ \alpha_Q \mathcal{L}_Q
+ w_H \mathcal{L}_H
+ \mathcal{L}_{reg}
+ w_R \mathcal{L}_R,
\end{equation}
where $k$ indexes the two BP targets, $w_k$ is the corresponding target weight, $\mathcal{L}_F$ and $\mathcal{L}_Q$ are focal-style and quantile residual penalties, $\mathcal{L}_H$ is an early-epoch Huber term, $\mathcal{L}_{reg}$ contains KAN and target-scale regularization, and $\mathcal{L}_R$ supervises the fused residual correction. For every mini-batch, the active terms are reduced to scalars and combined by weighted addition into the single objective $\mathcal{L}_{total}$ used for back-propagation. The word ``adaptive'' applies specifically to the target-wise squared-error weighting, not to every term in the sum. This target-weighting term is
\begin{equation}
\mathcal{L}_{A,k}=\frac{1}{2}\left[\exp(-s_k)\operatorname{MSE}_k+s_k+\log(2\pi)\right],
\end{equation}
where $s_k$ is a learnable target-specific scale parameter shared across all input windows. Together with the fixed target coefficient $w_k$, $\exp(-s_k)$ adjusts the relative contribution of the DBP and SBP squared errors during optimization. The coefficients $\alpha_F$ and $\alpha_Q$ follow an epoch-dependent ramp-up schedule, whereas $w_H=0.05$ for the first ten epochs and $w_H=0$ thereafter; the remaining regularization and residual-supervision coefficients are fixed during a run.

Each component addresses a different aspect of the regression problem:
\begin{itemize}
    \item \textbf{Adaptively weighted squared error ($\mathcal{L}_{A,k}$):} balances the two regression targets using learned global scale parameters.
    \item \textbf{Focal-style regression loss ($\mathcal{L}_F$):} emphasizes examples with large absolute errors.
    \item \textbf{Quantile residual loss ($\mathcal{L}_Q$):} introduces asymmetric penalties at $\tau=0.1$ and $\tau=0.9$ to improve robustness to uneven residual distributions.
    \item \textbf{Huber loss ($\mathcal{L}_H$):} provides additional robustness during the first ten training epochs.
    \item \textbf{Regularization and residual supervision ($\mathcal{L}_{reg}$ and $\mathcal{L}_R$):} regularize the KAN and target-scale parameters and align the fused correction with the observed residual.
\end{itemize}

\subsection{Implementation Details}
The proposed model was implemented in Python using PyTorch 2.10.0+cu128 and trained on a single NVIDIA GeForce RTX 4090 GPU. 
The model configuration and implementation settings are provided in Table~\ref{tab:hyperparams}.

\begin{table}[htbp]
  \centering
  \caption{Model configuration and implementation settings.}
  \label{tab:hyperparams}
  \small
  \setlength{\tabcolsep}{4pt}
  \renewcommand{\arraystretch}{1.05}
  \begin{tabularx}{\columnwidth}{p{0.45\columnwidth} X}
    \toprule
    \textbf{Parameter} & \textbf{Value} \\
    \midrule
    \textit{Input Configuration} & \\
    Source Waveform Sampling Rate & \SI{125}{Hz} \\
    Predictor Dimension ($F$) & 8 variables \\
    Model Sequence Length ($L$) & 10 feature records \\
    Window Step & 1 record \\
    Prediction Length & 1 record \\
    KAN Input Dimension & 80 \\
    XGBoost Input Dimension & 96 \\
    Batch Size ($B$) & 32 \\
    \midrule
    \textit{MSEM Encoder} & \\
    Transformer Layers & 6 \\
    Hidden Dimension ($d_{\mathrm{model}}$) & 128 \\
    Attention Heads & 8 \\
    \midrule
    \textit{Optimization} & \\
    Optimizer & RAdam with Lookahead \\
    Initial Learning Rate & $1 \times 10^{-3}$ \\
    Warmup Epochs & 10 \\
    LR Scheduler & Cosine Annealing \\
    \midrule
    \multicolumn{2}{l}{\textit{Software and Hardware Environment}} \\
    Deep Learning Framework & PyTorch 2.10.0+cu128 \\
    GPU & NVIDIA GeForce RTX 4090 \\
    \bottomrule
  \end{tabularx}
\end{table}

Training was performed using the RAdam optimizer wrapped with Lookahead to improve convergence stability. 
A cosine annealing learning rate schedule was adopted after a 10-epoch linear warmup phase.

\section{Experiments}
\subsection{Experimental Protocol}
\label{subsec:experimental_protocol}

All comparison and ablation experiments were conducted on the same model-ready table of 53,621 complete observations derived from the MIMIC dataset. 
To ensure a fair evaluation, all models used the same model-ready feature table, segment partition, training-only scaling procedure, sequence length, and target definitions for DBP and SBP prediction. 
Contiguous observations were first grouped into segments from discontinuities in the retained segment index. These segments were then randomly allocated to training, validation, and test partitions in an 80\%/10\%/10\% ratio using a single random seed, and the same split was used for all baseline, comparison, and ablation experiments unless otherwise specified. This procedure prevented windows derived from the same detected segment from crossing partitions. However, because the exported model-ready table did not retain the MIMIC subject identifier, the split was not subject-disjoint and different segments from the same subject may occur in more than one partition. Accordingly, ``held-out test set'' in this work refers to a segment-level holdout. All reported results correspond to a single seeded run; repeated-run variability was not evaluated.

For neural-network-based models, training was performed with early stopping on the validation set. 
Unless otherwise specified, the patience value was set to 20, and the checkpoint with the best monitored validation loss or validation metric was selected for final testing. 
For tree-based models such as XGBoost, the model was trained on the same training split and evaluated on the same test split. 
All reported results were computed on the segment-level held-out test set.

The comparison experiments were designed to evaluate the proposed framework against representative baseline models, including conventional deep learning architectures and standalone statistical or neural regressors. 
The ablation experiments were designed to isolate the contribution of major components in the proposed framework, including the XGBoost baseline/prior branch and the KAN branch.

\subsection{Evaluation Metrics}
\label{subsec:evaluation_metrics}

The model performance was evaluated using multiple error and clinical accuracy metrics. 
For each blood pressure target, the prediction error was defined as
\begin{equation}
    e_i = \hat{y}_i - y_i,
\end{equation}
where $\hat{y}_i$ denotes the predicted BP value and $y_i$ denotes the corresponding reference value.

The \textbf{mean error} was used to measure systematic bias:
\begin{equation}
    \mathrm{Mean} = \frac{1}{N}\sum_{i=1}^{N} e_i .
\end{equation}
A mean error closer to zero indicates lower systematic bias. 
The \textbf{standard deviation} (SD) of the errors was used to quantify the dispersion of prediction errors:
\begin{equation}
    \mathrm{SD} = 
    \sqrt{
    \frac{1}{N-1}
    \sum_{i=1}^{N}
    (e_i-\bar{e})^2
    },
\end{equation}
where $\bar{e}$ is the mean error. 
A smaller SD indicates more stable predictions.

We also reported the 95\% limits of agreement (LoA) based on Bland--Altman analysis:
\begin{equation}
    \mathrm{LoA}_{95\%} =
    \left[
    \bar{e} - 1.96 \times \mathrm{SD},
    \bar{e} + 1.96 \times \mathrm{SD}
    \right].
\end{equation}
A narrower LoA interval indicates better agreement between the predicted and reference BP values.
The reported bias, SD, and LoA are descriptive per-window point estimates for the 2,431 test windows. Because temporally adjacent windows and multiple segments can arise from the same subject, these observations are not statistically independent. Subject-clustered confidence intervals and repeated-measures Bland--Altman analysis were not calculated because subject identifiers were not retained in the evaluation export. The reported LoA should therefore not be interpreted as subject-level precision and may be narrower than estimates obtained with subject-clustered inference.

In addition, cumulative error percentages were calculated at the \SI{5}{mmHg}, \SI{10}{mmHg}, and \SI{15}{mmHg} thresholds:
\begin{equation}
    P_{\leq T} =
    \frac{1}{N}
    \sum_{i=1}^{N}
    \mathbb{I}(|e_i| \leq T)
    \times 100\% ,
\end{equation}
where $T \in \{5,10,15\}$ mmHg and $\mathbb{I}(\cdot)$ is the indicator function. 
These cumulative percentages were further compared with the numerical percentage thresholds defined by the British Hypertension Society (BHS).
These numerical comparisons do not constitute formal compliance testing, clinical device validation, or certification under ISO 81060-3/AAMI or the BHS protocol.

\subsection{Comparison with Representative Baseline Models}
\label{subsec:comparison}

To evaluate the proposed framework against representative architectures, we selected six baseline families with reference to their established use in cuffless BP estimation or their foundational formulations: CNN--LSTM \citep{jeong2021cnnlstm}, ResNet-1D \citep{liu2025resnetbilstm}, U-Net 1D \citep{athaya2021unet}, XGBoost \citep{chen2016xgboost}, Transformer \citep{vaswani2017attention}, and KAN \citep{liu2024kan}.
The XGBoost baseline uses flattened statistical features for direct regression, the Transformer baseline uses a temporal neural backbone, and the KAN baseline uses a standalone nonlinear neural regressor.
These references identify the corresponding architecture families; the numerical results in Table~\ref{tab:sota_bhs} were not taken from the cited studies. All six baselines were implemented and retrained locally, then evaluated using the same model-ready inputs, data split, preprocessing protocol, and target definitions as the proposed framework.

\begin{table*}[htbp]
  \centering
  \caption{Comparison with representative baseline models. Arrows indicate whether lower ($\downarrow$) or higher ($\uparrow$) values are better. The descriptive per-window 95\% limits of agreement (LoA) are calculated as Mean $\pm$ 1.96 SD without repeated-measures adjustment.}
  \label{tab:sota_bhs}
  \setlength{\tabcolsep}{2.2pt}
  \scriptsize
  \resizebox{\textwidth}{!}{%
  \begin{tabular}{l|ccc|ccc|ccc|ccc}
    \toprule
    \multirow{3}{*}{\textbf{Model}} 
    & \multicolumn{6}{c|}{\textbf{DBP (mmHg)}} 
    & \multicolumn{6}{c}{\textbf{SBP (mmHg)}} \\
    \cmidrule(lr){2-7} \cmidrule(lr){8-13}
    & \multicolumn{3}{c|}{\textbf{Error Agreement}} 
    & \multicolumn{3}{c|}{\textbf{Cumulative Accuracy (\%)}} 
    & \multicolumn{3}{c|}{\textbf{Error Agreement}} 
    & \multicolumn{3}{c}{\textbf{Cumulative Accuracy (\%)}} \\
    & Mean$^{\dagger}$ & SD $\downarrow$ & 95\% LoA
    & $\le 5 \uparrow$ & $\le 10 \uparrow$ & $\le 15 \uparrow$ 
    & Mean$^{\dagger}$ & SD $\downarrow$ & 95\% LoA
    & $\le 5 \uparrow$ & $\le 10 \uparrow$ & $\le 15 \uparrow$ \\
    \midrule
    CNN-LSTM 
    & 0.49 & 4.54 & [-8.41, 9.39] & \textbf{88.21} & 95.46 & 97.73
    & 0.94 & 7.61 & [-13.98, 15.86] & 67.57 & 85.11 & 92.06 \\

    ResNet-1D 
    & 0.50 & 5.06 & [-9.42, 10.42] & 82.01 & 93.73 & 96.22
    & 1.09 & 8.13 & [-14.84, 17.02] & 64.78 & 83.90 & 91.38 \\

    U-Net 1D 
    & -0.41 & 4.55 & [-9.33, 8.51] & \textbf{88.21} & 94.94 & 97.20
    & 0.67 & 8.45 & [-15.89, 17.23] & 65.84 & 86.09 & 91.91 \\

    \midrule
    XGBoost 
    & 0.41 & 3.86 & [-7.16, 7.98] & 87.53 & 97.43 & 99.01
    & \textbf{-0.60} & 7.10 & [-14.52, 13.32] & 64.78 & 87.23 & 93.73 \\

    Transformer 
    & -0.75 & 4.76 & [-10.08, 8.58] & 84.88 & 96.90 & 98.34
    & 0.91 & 7.80 & [-14.38, 16.20] & 64.85 & 84.13 & 91.16 \\

    KAN 
    & -0.91 & 4.13 & [-9.00, 7.18] & 83.26 & 98.03 & \textbf{99.05}
    & -3.55 & 8.57 & [-20.35, 13.25] & 53.15 & 75.65 & 93.09 \\

    \midrule
    \textbf{Proposed} 
    & 0.41 & \textbf{3.74} & \textbf{[-6.93, 7.74]} & 87.95 & \textbf{98.48} & 99.01
    & -1.60 & \textbf{5.95} & \textbf{[-13.25, 10.06]} & \textbf{70.75} & \textbf{94.36} & \textbf{98.11} \\
    \bottomrule
    \multicolumn{13}{l}{\footnotesize $^{\dagger}$ For Mean error, a value closer to zero indicates lower systematic bias. For 95\% LoA, a narrower interval indicates better agreement.}
  \end{tabular}%
  }
\end{table*}

Table~\ref{tab:sota_bhs} reports the comparison results. 
Among all compared methods, the proposed model achieves the best overall DBP agreement, with the lowest DBP SD of \SI{3.74}{mmHg} and the narrowest DBP 95\% LoA of [-6.93, 7.74] mmHg. Its cumulative accuracies are 87.95\%, 98.48\%, and 99.01\% at the \SI{5}{mmHg}, \SI{10}{mmHg}, and \SI{15}{mmHg} thresholds, respectively. The \SI{10}{mmHg} accuracy is the highest among the compared methods, while the \SI{5}{mmHg} and \SI{15}{mmHg} accuracies remain within 0.26 and 0.04 percentage points of the corresponding best results. 
These results indicate that the proposed framework provides the strongest DBP agreement while maintaining highly competitive threshold-based accuracy.

For SBP estimation, XGBoost obtains the mean error closest to zero, with a bias of -\SI{0.60}{mmHg}. 
However, its SBP SD and LoA remain larger than those of the proposed model. 
The proposed model achieves the lowest SBP SD of \SI{5.95}{mmHg}, the narrowest SBP 95\% LoA of [-13.25, 10.06] mmHg, and the highest SBP cumulative accuracies of 70.75\%, 94.36\%, and 98.11\% at the three error thresholds. 
This suggests that the proposed framework provides better agreement and stability for SBP prediction, even though its mean error is not the closest to zero.

The XGBoost baseline remains a strong statistical baseline, especially compared with standalone neural architectures. 
Nevertheless, the proposed model outperforms XGBoost in DBP and SBP SD and 95\% LoA, while matching or improving its cumulative accuracy at all thresholds. 
Compared with XGBoost, the proposed model reduces the SBP SD from \SI{7.10}{mmHg} to \SI{5.95}{mmHg}, narrows the SBP LoA from [-14.52, 13.32] mmHg to [-13.25, 10.06] mmHg, and improves the SBP $\le\SI{10}{mmHg}$ accuracy from 87.23\% to 94.36\%. 
These improvements indicate that the hybrid design does not merely rely on the XGBoost prior, but further refines it through neural residual correction.

The KAN baseline achieves reasonable DBP performance, but its SBP performance is weaker, with a wider SBP LoA of [-20.35, 13.25] mmHg and lower cumulative accuracy. 
This suggests that KAN alone has nonlinear regression capability but is insufficient for stable SBP prediction when used as an independent predictor. 
Overall, the proposed model provides the most balanced performance across DBP and SBP by combining statistical prior estimation with conservative neural refinement.

\textbf{Key Observations:}
\begin{itemize}
    \item \textbf{Improved agreement:} The proposed model achieves the narrowest DBP and SBP LoA intervals, indicating better agreement with the reference BP values.

    \item \textbf{Stronger SBP stability:} Compared with XGBoost, the proposed framework reduces SBP SD and substantially improves SBP cumulative accuracy, especially at the $\le\SI{10}{mmHg}$ threshold.

    \item \textbf{Balanced hybrid prediction:} Standalone neural models and single-branch baselines show less consistent behavior across DBP and SBP, whereas the proposed model maintains strong performance for both targets.
\end{itemize}

\subsection{Ablation Study}
\label{subsec:ablation}

To assess the conditional contribution of each major component within the complete framework, we used a leave-one-component-out (LOCO) ablation design. Starting from the full model, each reduced variant removes exactly one component while retaining the other two components and the same data split, preprocessing, target definitions, and training protocol. The \textbf{Hybrid w/o KAN} variant removes the KAN branch while retaining the XGBoost baseline/prior branch and MHDA-based fusion; the \textbf{Hybrid w/o XGB} variant removes the XGBoost branch while retaining KAN and MHDA; and the \textbf{Hybrid w/o MHDA} variant removes MHDA while retaining the XGBoost and KAN branches. The \textbf{Proposed} model contains all three components. This LOCO design evaluates the marginal degradation caused by removing a component from the complete system; it is not a sequential-addition or full-factorial analysis of all possible module interactions.

\begin{table*}[htbp]
  \centering
  \caption{Leave-one-component-out ablation study of the proposed hybrid baseline-residual framework. A filled circle indicates that a component is retained, whereas an en dash indicates that it is removed. Arrows indicate whether lower ($\downarrow$) or higher ($\uparrow$) values are better. The descriptive per-window 95\% limits of agreement (LoA) are calculated as Mean $\pm$ 1.96 SD without repeated-measures adjustment.}
  \label{tab:ablation}
  \setlength{\tabcolsep}{2.0pt}
  \scriptsize
  \resizebox{\textwidth}{!}{%
  \begin{tabular}{l|ccc|ccc|ccc|ccc|ccc}
    \toprule
    \multirow{3}{*}{\textbf{Configuration}} 
    & \multicolumn{3}{c|}{\textbf{Components retained}}
    & \multicolumn{6}{c|}{\textbf{DBP (mmHg)}} 
    & \multicolumn{6}{c}{\textbf{SBP (mmHg)}} \\
    \cmidrule(lr){2-4} \cmidrule(lr){5-10} \cmidrule(lr){11-16}
    & \multirow{2}{*}{\textbf{XGB}} & \multirow{2}{*}{\textbf{KAN}} & \multirow{2}{*}{\textbf{MHDA}}
    & \multicolumn{3}{c|}{\textbf{Error Agreement}} 
    & \multicolumn{3}{c|}{\textbf{Cumulative Accuracy (\%)}} 
    & \multicolumn{3}{c|}{\textbf{Error Agreement}} 
    & \multicolumn{3}{c}{\textbf{Cumulative Accuracy (\%)}} \\
    & & &
    & Mean$^{\dagger}$ & SD $\downarrow$ & 95\% LoA
    & $\le 5 \uparrow$ & $\le 10 \uparrow$ & $\le 15 \uparrow$ 
    & Mean$^{\dagger}$ & SD $\downarrow$ & 95\% LoA
    & $\le 5 \uparrow$ & $\le 10 \uparrow$ & $\le 15 \uparrow$ \\
    \midrule
    Hybrid w/o KAN
    & $\bullet$ & -- & $\bullet$
    & 2.26 & 4.70 & [-6.95, 11.47] & 67.46 & 95.48 & 99.18
    & 7.26 & 15.15 & [-22.43, 36.95] & 26.74 & 40.07 & 56.56 \\

    Hybrid w/o XGB
    & -- & $\bullet$ & $\bullet$
    & 2.63 & 12.00 & [-20.89, 26.15] & 19.09 & 51.21 & 68.74
    & -4.29 & 23.43 & [-50.21, 41.63] & 13.74 & 31.18 & 41.83 \\

    Hybrid w/o MHDA
    & $\bullet$ & $\bullet$ & --
    & 2.84 & 4.79 & [-6.55, 12.23] & 65.49 & 92.80 & \textbf{99.30}
    & 9.95 & 13.94 & [-17.37, 37.27] & 26.12 & 42.62 & 59.11 \\

    \textbf{Proposed}
    & $\bullet$ & $\bullet$ & $\bullet$
    & \textbf{0.41} & \textbf{3.74} & \textbf{[-6.93, 7.74]} & \textbf{87.95} & \textbf{98.48} & 99.01
    & \textbf{-1.60} & \textbf{5.95} & \textbf{[-13.25, 10.06]} & \textbf{70.75} & \textbf{94.36} & \textbf{98.11} \\
    \bottomrule
    \multicolumn{16}{l}{\footnotesize $^{\dagger}$ For Mean error, a value closer to zero indicates lower systematic bias. For 95\% LoA, a narrower interval indicates better agreement.}
  \end{tabular}%
  }
\end{table*}

Table~\ref{tab:ablation} summarizes the ablation results. 
Removing the XGBoost baseline/prior branch causes the most severe degradation. 
Relative to the full model, the \textbf{Hybrid w/o XGB} variant increases DBP SD by \SI{8.26}{mmHg}, from 3.74 to \SI{12.00}{mmHg}, and SBP SD by \SI{17.48}{mmHg}, from 5.95 to \SI{23.43}{mmHg}. 
The corresponding 95\% LoA intervals become substantially wider, reaching [-20.89, 26.15] mmHg for DBP and [-50.21, 41.63] mmHg for SBP. 
This large expansion of the LoA intervals indicates poor agreement with the reference BP values and supports the role of the XGBoost branch as a statistical anchor in the hybrid residual framework.

The \textbf{Hybrid w/o KAN} variant also shows clear performance degradation, especially for SBP. 
Although the XGBoost baseline/prior branch is retained, removing KAN increases SBP SD by \SI{9.20}{mmHg}, widens the SBP LoA to [-22.43, 36.95] mmHg, and reduces the SBP $\le\SI{10}{mmHg}$ accuracy by 54.29 percentage points, from 94.36\% to 40.07\%. 
This suggests that the KAN branch contributes to nonlinear physiological approximation and helps stabilize the residual correction process, particularly for SBP estimation.

The \textbf{Hybrid w/o MHDA} variant further demonstrates the importance of the MHDA module. 
After removing MHDA, the DBP SD increases by \SI{1.05}{mmHg}, from 3.74 to \SI{4.79}{mmHg}, and the DBP $\le\SI{10}{mmHg}$ accuracy decreases by 5.68 percentage points, from 98.48\% to 92.80\%. 
The degradation is more pronounced for SBP: the mean error increases to \SI{9.95}{mmHg}, the SD increases to \SI{13.94}{mmHg}, and the 95\% LoA widens to [-17.37, 37.27] mmHg. 
The SBP $\le\SI{10}{mmHg}$ accuracy also decreases markedly from 94.36\% to 42.62\%. 
These results indicate that MHDA plays an important role in temporal feature interaction and in suppressing unstable residual corrections, especially for SBP estimation.

Compared with the reduced variants, the \textbf{Proposed} model achieves the narrowest LoA intervals for both DBP and SBP, with [-6.93, 7.74] mmHg for DBP and [-13.25, 10.06] mmHg for SBP. 
It also achieves the best SBP cumulative accuracy at all three thresholds and the best DBP $\le\SI{5}{mmHg}$ and $\le\SI{10}{mmHg}$ accuracies. 
Although the \textbf{Hybrid w/o MHDA} variant has a slightly higher DBP $\le\SI{15}{mmHg}$ accuracy, its bias, SD, LoA, and SBP performance are substantially worse. 
Therefore, the full framework provides the most stable overall performance by combining the XGBoost statistical anchor, KAN-assisted nonlinear modeling, MHDA-based feature interaction, and conservative residual refinement.

\textbf{Analysis of Results:}
\begin{itemize}
    \item \textbf{Role of XGBoost:} Removing the XGBoost branch leads to the widest LoA intervals and the largest drop in cumulative accuracy, indicating that the XGBoost prior is important for stabilizing the hybrid prediction framework.

    \item \textbf{Role of KAN:} Removing KAN substantially degrades SBP performance, suggesting that KAN helps model nonlinear physiological patterns and prevents unstable residual correction.

    \item \textbf{Role of MHDA:} Removing MHDA causes clear degradation in both DBP and SBP, especially in SBP SD, LoA, and threshold accuracy, indicating that MHDA is important for effective temporal feature interaction and robust residual correction.

    \item \textbf{Overall stability:} The proposed model does not achieve the best value in every single threshold-related column, but it provides the best overall agreement-oriented performance, especially in SD, LoA, and SBP cumulative accuracy.
\end{itemize}

\subsection{Statistical Analysis and Visualization}

To comprehensively evaluate the model's reliability and dynamic tracking capability, we performed a detailed visualization analysis using the segment-level held-out test set and a separate extended held-out visualization set.

\begin{figure*}[t!]
    \centering
    \includegraphics[width=\linewidth]{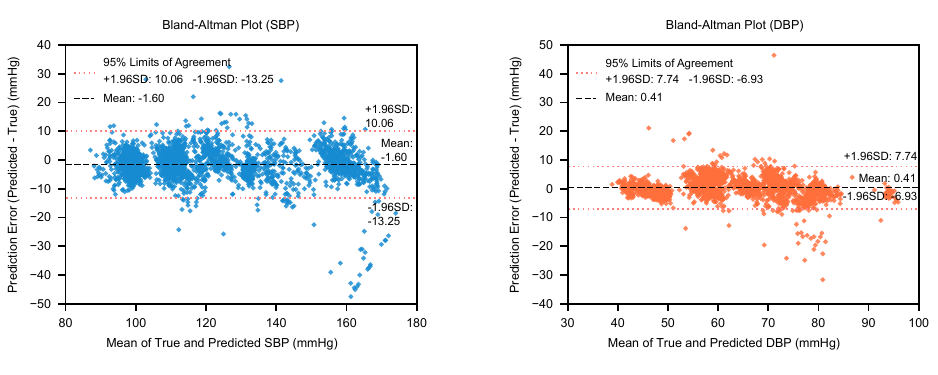}
    
    \vspace{1em} 
    

        \centering
{\includegraphics[width=\linewidth]{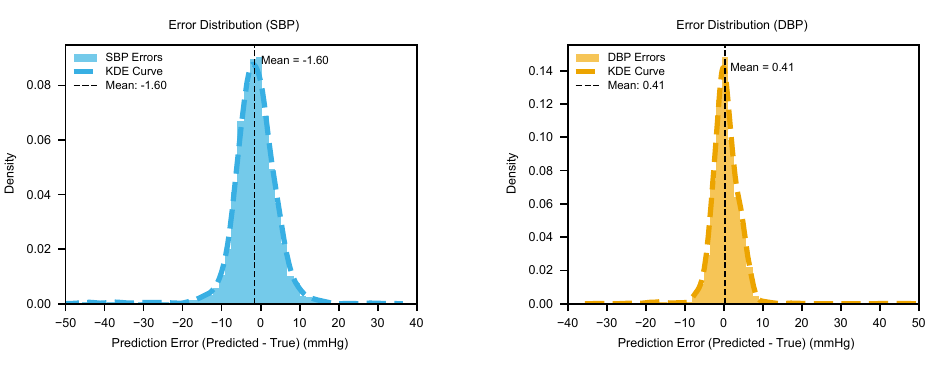}}
    
    \vspace{1em}
    

        \centering
        \includegraphics[width=\linewidth]{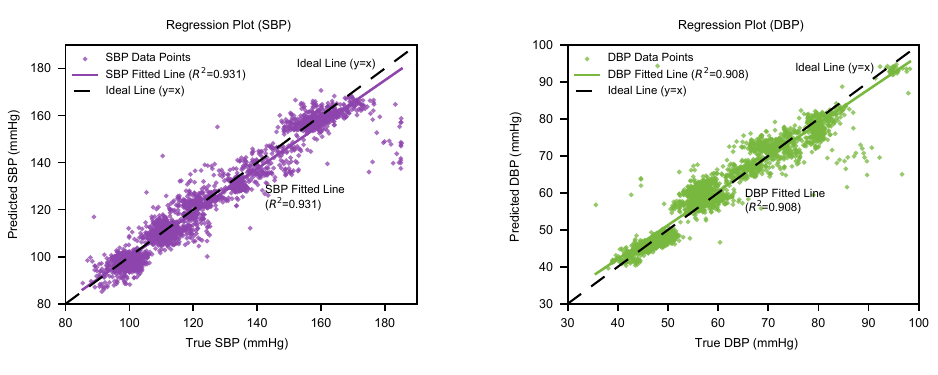}
    \caption{\textbf{Visualization of prediction performance.}
    The top row presents descriptive per-window Bland--Altman plots showing agreement between predicted and reference BP. The dotted red lines represent the unadjusted 95\% limits of agreement ($\pm 1.96$ SD), and the dashed black line represents the mean bias; repeated measurements within subjects are not accounted for.
    The middle row presents error-distribution histograms with kernel density estimation (KDE) curves, showing that the prediction errors remain concentrated near zero.
    The bottom row presents scatter plots with high coefficients of determination ($R^2 > 0.90$) between the estimated and reference values.}
    \label{fig:visualization}
\end{figure*}

\textbf{Agreement and Error Distribution:} 
Fig.~\ref{fig:visualization} presents the descriptive per-window agreement analysis. The Bland--Altman plots in the top row show mean biases of \SI{0.41}{mmHg} for DBP and \SI{-1.60}{mmHg} for SBP, with unadjusted 95\% LoA intervals of [-6.93, 7.74] mmHg and [-13.25, 10.06] mmHg, respectively. The error distributions in the middle row remain concentrated near zero with limited tails, consistent with the small overall biases. Furthermore, the regression scatter plots in the bottom row have high coefficients of determination ($R^2 > 0.90$) between the estimated and reference values across the evaluated range. These plots do not account for repeated measurements within subjects.

\begin{figure*}[t!]
    \centering
    \includegraphics[width=\textwidth]{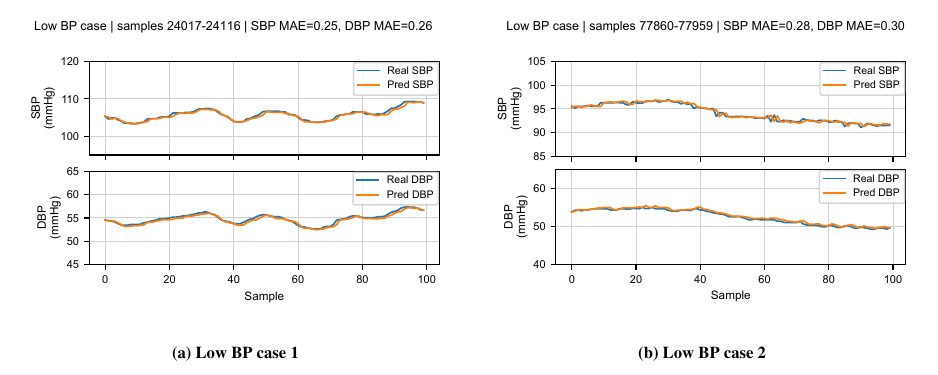}
    \vspace{0.8em}
    \includegraphics[width=\textwidth]{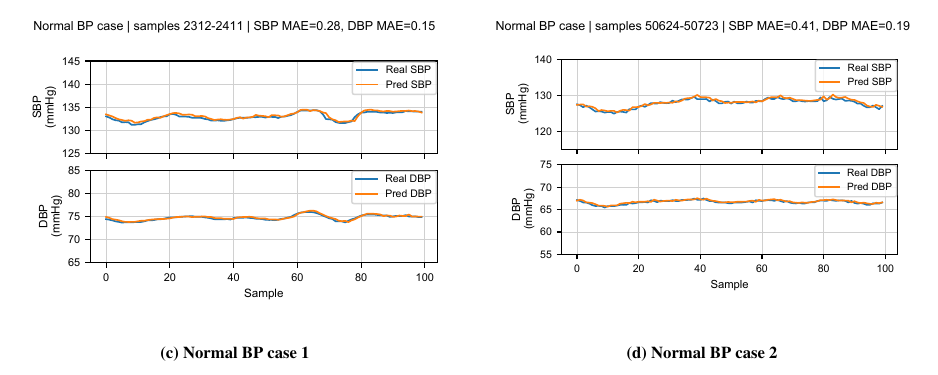}
    \vspace{0.8em}
    \includegraphics[width=\textwidth]{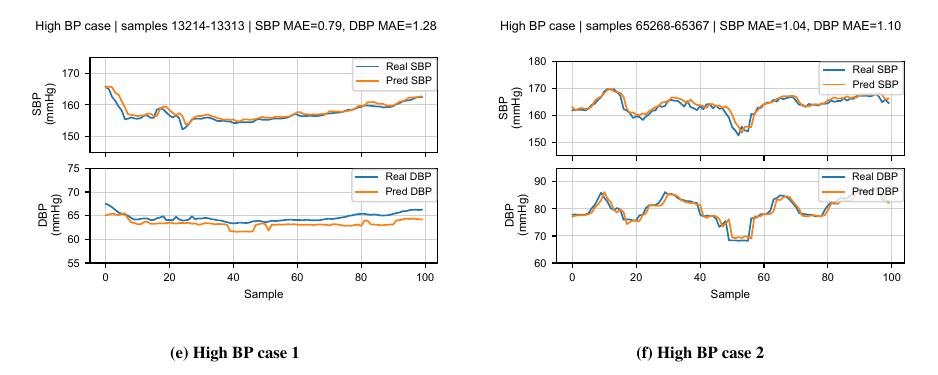}
    \caption{\textbf{Representative continuous blood pressure tracking cases under different BP ranges.}
    Six locally zoomed segments from an extended held-out visualization set are shown, including two low-BP cases, two normal-BP cases, and two high-BP cases. This set is separate from the test set summarized in Table~\ref{tab:sota_bhs}, while the same trained model is used.
    Each case separately presents SBP and DBP trajectories within a local coordinate range, making trajectory-level agreement and short-term prediction errors easier to inspect.
    The predicted curves remain closely aligned with the reference trajectories across the selected low, normal, and high BP cases, illustrating consistent tracking behavior across different pressure ranges.}
    \label{fig:tracking}
\end{figure*}

\textbf{Representative Continuous Tracking Cases:}
Beyond static statistical metrics, the ability to continuously follow blood pressure variations under different physiological ranges is essential for real-time monitoring. 
To provide a clearer visual assessment, Fig.~\ref{fig:tracking} presents six representative segments from an extended held-out visualization set evaluated with the same trained model used for Table~\ref{tab:sota_bhs}. The segments cover low, normal, and high blood-pressure ranges, with two examples shown for each range. 
Compared with plotting a long and dense sequence, these locally zoomed segments allow the agreement between reference and predicted SBP/DBP trajectories to be inspected more clearly.

As shown in Fig.~\ref{fig:tracking}, the proposed model closely follows the reference BP trajectories across different pressure levels. 
In the low-pressure cases, the model tracks both gradual trends and local fluctuations.
In the normal-pressure cases, the predicted curves remain stable and well aligned with the reference values. 
In the high-pressure cases, the model maintains accurate tracking under elevated SBP/DBP ranges, indicating that the proposed framework is capable of capturing BP dynamics across a broad physiological spectrum.

\clearpage
\subsection{Numerical Comparison with BHS Thresholds}
Finally, we compared the cumulative error percentages of the proposed method with the numerical thresholds defined by the British Hypertension Society (BHS).
Table~\ref{tab:bhs} separates the BHS reference percentages from the two outputs of the proposed model: the first row lists the minimum cumulative percentages associated with the numerical Grade~A thresholds, whereas the subsequent rows report the DBP and SBP results produced by the same proposed model on the segment-level test set.

For DBP, the cumulative error percentages reach 87.95\%, 98.48\%, and 99.01\% at the \SI{5}{mmHg}, \SI{10}{mmHg}, and \SI{15}{mmHg} thresholds, respectively, all of which exceed the numerical Grade A percentage thresholds. 
For SBP, the model achieves 70.75\%, 94.36\%, and 98.11\% at the same three thresholds, likewise falling within these numerical thresholds. 

Thus, both BP targets exceed all three reference percentages. This is a descriptive threshold comparison rather than a formal BHS grade assignment because the evaluation did not follow a protocol-specific clinical validation design.

\begin{table}[H]
  \centering
  \caption{BHS Grade~A reference percentages and the corresponding cumulative error percentages of the proposed model for DBP and SBP. This descriptive comparison is not a formal BHS grade assignment.}
  \label{tab:bhs}
  \small
  \setlength{\tabcolsep}{3pt}
  \renewcommand{\arraystretch}{1.08}
  \begin{tabularx}{\columnwidth}{@{}>{\raggedright\arraybackslash}p{0.19\columnwidth}>{\centering\arraybackslash}p{0.11\columnwidth} *{3}{>{\centering\arraybackslash}X} >{\centering\arraybackslash}p{0.19\columnwidth}@{}}
    \toprule
    \multirow{2}{=}{\textbf{Row type}} & \multirow{2}{=}{\centering\textbf{BP target}} & \multicolumn{3}{c}{\textbf{Cumulative Error Percentage (\%)}} & \multirow{2}{=}{\centering\textbf{Numerical comparison}} \\
    & & $\le$\SI{5}{mmHg} & $\le$\SI{10}{mmHg} & $\le$\SI{15}{mmHg} & \\
    \midrule
    BHS Grade~A reference & -- & 60.00 & 85.00 & 95.00 & Reference threshold \\
    \midrule
    \multirow{2}{=}{Proposed model output} & \textbf{DBP} & 87.95 & 98.48 & 99.01 & All three met \\
    & \textbf{SBP} & 70.75 & 94.36 & 98.11 & All three met \\
    \bottomrule
  \end{tabularx}
\end{table}

\section{Discussion}

\subsection{Interpretation of Performance}
The experimental results demonstrate that the proposed hybrid framework achieves the best overall agreement-oriented performance among the compared models for continuous blood pressure estimation. 
Unlike end-to-end waveform models (e.g., ResNet-1D) that rely solely on latent feature extraction, our method explicitly integrates six ECG/PPG-derived descriptors (PTT, heart rate, foot-to-peak interval, maximum PPG slope, RI, and SD time) with age and body weight. The ablation study confirms that the inclusion of the XGBoost branch and KAN network significantly reduces prediction bias, particularly for SBP, which typically exhibits higher variability than DBP.

\subsection{Clinical Implications}
From a clinical perspective, the model's values fall within the numerical BHS Grade A percentage thresholds and numerical AAMI mean-error and SD criteria on the segment-level held-out test set, motivating formal clinical validation. These retrospective numerical comparisons do not constitute compliance testing, device validation, or certification. The tracking behavior observed in the selected hypotensive segments (Fig.~\ref{fig:tracking}) further motivates evaluation in ICU monitoring and wearable health settings.

\subsection{Limitations and Future Work}
Despite the promising results, this study has limitations. 
First, the model was validated on the MIMIC-III database, which consists of data from ICU patients. The signal quality in ambulatory settings (e.g., daily activities) may be lower due to motion artifacts, and the model's robustness in such scenarios requires further validation.
Second, although we used calibration-free features, adding periodic calibration could further enhance long-term accuracy. Future work will focus on compressing the model for deployment on edge devices (e.g., smartwatches) and validating it on real-world wearable datasets.

Third, the present evaluation retained the original segment-level split because subject identifiers were not available when the model-ready feature table and experimental partitions were created. Subject mapping was subsequently reconstructed from the source metadata and waveform archive for cohort accounting, but it was not used to retrospectively alter the established partitions or to claim subject-disjoint performance. Although windows from an individual segment were confined to one partition, segments from the same subject may therefore have appeared in different partitions. This can yield more optimistic estimates of generalization than a strictly subject-disjoint evaluation. Future work should repeat model development and evaluation using a de novo subject-disjoint split defined before scaling and window construction, together with external-cohort validation.

Fourth, the original complete-case pipeline did not apply pre-specified physiological-range exclusions or RI clipping. Although the held-out test references contained no $\mathrm{SBP}\leq\mathrm{DBP}$ rows, the influence of the extreme values retained elsewhere in the model-ready table on training was not quantified through a retrained sensitivity analysis. Future evaluation should predefine physiologically justified quality-control rules and compare results with and without those exclusions.

Fifth, the 2,431 test windows include temporally adjacent and repeated observations from some subjects. The reported SD and Bland--Altman LoA treat windows as individual observations and do not include subject-clustered confidence intervals or a repeated-measures Bland--Altman model. They should therefore be interpreted as descriptive per-window agreement estimates. Subject-clustered bootstrap inference or repeated-measures agreement analysis should accompany the de novo subject-disjoint evaluation.

\section{Conclusion}
In this paper, we proposed a multi-source data-driven framework for cuffless continuous blood pressure monitoring. The method models standardized 10-step sequences of six ECG/PPG-derived physiological descriptors and two demographic covariates using a hybrid Transformer architecture (MSEM and DCFD), while XGBoost and KAN provide complementary tabular and nonlinear representations. Experimental results on the MIMIC-III dataset show that our model achieves the best overall agreement-oriented performance among the compared baselines and falls within the numerical AAMI mean-error and SD criteria and numerical BHS Grade A percentage thresholds on the segment-level test set. These numerical comparisons are descriptive and do not constitute formal compliance testing or device certification. The ablation study and visualization further demonstrate the contribution of the model components and its dynamic tracking behavior on the evaluated data. These findings should be confirmed using pre-specified quality-control rules, repeated-measures inference, de novo subject-disjoint evaluation, and an external cohort before clinical generalization is inferred.

\section*{Ethical statement}
This study was a secondary analysis of de-identified physiological and linked demographic data from the MIMIC-III Waveform Database and MIMIC-III Clinical Database. No participants were recruited and no new physiological or clinical data were collected by the authors. The secondary analysis was conducted in accordance with the principles of the Declaration of Helsinki and the applicable PhysioNet credentialing and data-use requirements. According to the database documentation, the original MIMIC-III project was approved by the Institutional Review Boards of Beth Israel Deaconess Medical Center and the Massachusetts Institute of Technology, and the requirement for individual patient consent was waived because the project did not affect clinical care and all protected health information was de-identified \citep{johnson2016mimicdb,johnson2016mimic}.

\funding{This research received no external funding. The costs of the computational resources used in this study were supported by Zhaoying Liu.}

\section*{Acknowledgements}
The authors used OpenAI Codex (GPT-5.6) to assist with English-language editing, manuscript organization, LaTeX formatting, and literature-search support. All AI-assisted content was critically reviewed and verified by the authors, who take full responsibility for the accuracy, integrity, and originality of the manuscript.

\data{The MIMIC-III Waveform Database used in this study is available from PhysioNet (doi: \url{https://doi.org/10.13026/c2607m}) \citep{moody2020mimic3wdb}. The linked clinical and demographic data from the MIMIC-III Clinical Database are available from PhysioNet (doi: \url{https://doi.org/10.13026/C2XW26}) to credentialed users who complete the required training and accept the applicable data use agreement \citep{johnson2016mimicdb}. The authors are not permitted to redistribute restricted source records. The analysis code and non-identifying derived experimental data supporting the tables and figures are available from the corresponding author upon reasonable request, where disclosure is permitted by institutional requirements and the applicable PhysioNet data-use terms.}

\section*{Conflict of interest}
The authors declare no competing interests.

\roles{
Yuexin Ma: Conceptualization, Methodology, Investigation, Project administration, and Writing--review and editing. Yuexin Ma conceived the study, designed the initial model architecture and methodological innovations, established the evaluation framework, led the literature investigation, and coordinated the research direction.

Jingqi Hou: Methodology, Software, Data curation, Investigation, Validation, Formal analysis, Visualization, Writing--original draft, and Writing--review and editing. Jingqi Hou constructed the model-ready dataset from the source data, conducted the main model training and performance optimization, performed the baseline comparisons and ablation experiments, analyzed the results, prepared the figures, and led manuscript drafting, revision, and adaptation to the journal requirements.

Yuxuan Kang: Methodology, Software, Investigation, Validation, Visualization, and Writing--original draft. Yuxuan Kang contributed to exploratory evaluation of candidate model components, hyperparameter selection, figure preparation, and manuscript drafting.

Zhaoying Liu: Methodology, Validation, Supervision, Resources, and Writing--review and editing. Zhaoying Liu provided methodological guidance, critically evaluated the initial experimental evidence and manuscript, identified issues that motivated model refinement and the repetition of key experiments, guided interpretation of the revised results, provided computational resources, and critically revised the manuscript for important intellectual content.

All authors reviewed and approved the final manuscript.
}



\begin{thebibliography}{99}

\bibitem[Athaya and Choi(2021)]{athaya2021unet}
Athaya T and Choi S 2021 An estimation method of continuous non-invasive arterial blood pressure waveform using photoplethysmography: a U-Net architecture-based approach \emph{Sensors} \textbf{21} 1867 (doi: 10.3390/s21051867)

\bibitem[Athaya and Choi(2022)]{athaya2022squeezeunet}
Athaya T and Choi S 2022 Real-time cuffless continuous blood pressure estimation using 1D Squeeze U-Net model: a progress toward mHealth \emph{Biosensors} \textbf{12} 655 (doi: 10.3390/bios12080655)

\bibitem[Batra et al.(2026)]{batra2026ffdm}
Batra P, Maan P and Kumar M 2026 FFDM-GAT-transformer: a fast FDM-based graph-attention and transformer framework for cuffless blood pressure estimation using PPG signal \emph{Measurement} 121638 (doi: 10.1016/j.measurement.2026.121638)

\bibitem[Block et al.(2020)]{block2020cptt}
Block R C \emph{et al} 2020 Conventional pulse transit times as markers of blood pressure changes in humans \emph{Sci. Rep.} \textbf{10} 16373 (doi: 10.1038/s41598-020-73143-8)

\bibitem[Chang et al.(2024)]{chang2024tdstf}
Chang P \emph{et al} 2024 A transformer-based diffusion probabilistic model for heart rate and blood pressure forecasting in intensive care unit \emph{Comput. Methods Programs Biomed.} \textbf{246} 108060 (doi: 10.1016/j.cmpb.2024.108060)

\bibitem[Chen and Guestrin(2016)]{chen2016xgboost}
Chen T and Guestrin C 2016 XGBoost: a scalable tree boosting system \emph{Proc. 22nd ACM SIGKDD Int. Conf. on Knowledge Discovery and Data Mining} pp 785--94 (doi: 10.1145/2939672.2939785)

\bibitem[Chen et al.(2025)]{chen2025sixmodal}
Chen Y, Xu F, Huang Z, He J and Feng Z 2025 Cuffless blood pressure estimation from six wearable sensor modalities in multi-motion-state scenarios arXiv:2512.01653

\bibitem[Cheng et al.(2026)]{cheng2026ghcnet}
Cheng Y, Li T, Zhang Z, Huang Z, Mei Z and Vai M I 2026 GHC-net: a Gramian angular field based hybrid CNN for cuffless blood pressure classification using PPG signals \emph{Comput. Biol. Med.} \textbf{206} 111622 (doi: 10.1016/j.compbiomed.2026.111622)

\bibitem[Cisnal et al.(2026)]{cisnal2026trustworthy}
Cisnal A, Podder I, Grossmann L, Dheman K, Elgendi M, Valgimigli M and Paez-Granados D 2026 Towards trustworthy AI-driven cuffless blood pressure monitoring \emph{npj Digit. Med.} (doi: 10.1038/s41746-026-02898-7)

\bibitem[Colvonen(2021)]{colvonen2021response}
Colvonen P J 2021 Response to: Investigating sources of inaccuracy in wearable optical heart rate sensors \emph{npj Digit. Med.} \textbf{4} 38 (doi: 10.1038/s41746-021-00408-5)

\bibitem[Ding et al.(2016)]{ding2016pir}
Ding X-R, Zhang Y-T, Liu J, Dai W-X and Tsang H K 2016 Continuous cuffless blood pressure estimation using pulse transit time and photoplethysmogram intensity ratio \emph{IEEE Trans. Biomed. Eng.} \textbf{63} 964--72 (doi: 10.1109/TBME.2015.2480679)

\bibitem[Duan et al.(2024)]{duan2024fusion}
Duan J, Xiong J, Li Y and Ding W 2024 Deep learning based multimodal biomedical data fusion: an overview and comparative review \emph{Inf. Fusion} \textbf{112} 102536 (doi: 10.1016/j.inffus.2024.102536)

\bibitem[Heidari et al.(2016)]{heidari2016sager}
Heidari S, Babor T F, De Castro P, Tort S and Curno M 2016 Sex and Gender Equity in Research: rationale for the SAGER guidelines and recommended use \emph{Res. Integr. Peer Rev.} \textbf{1} 2 (doi: 10.1186/s41073-016-0007-6)

\bibitem[Hua et al.(2024)]{hua2024wearable}
Hua J \emph{et al} 2024 Wearable cuffless blood pressure monitoring: from flexible electronics to machine learning \emph{Wearable Electron.} \textbf{1} 78--90 (doi: 10.1016/j.wees.2024.05.004)

\bibitem[Huber(1964)]{huber1964}
Huber P J 1964 Robust estimation of a location parameter \emph{Ann. Math. Stat.} \textbf{35} 73--101 (doi: 10.1214/aoms/1177703732)

\bibitem[Ibtehaz et al.(2022)]{ibtehaz2022ppg2abp}
Ibtehaz N \emph{et al} 2022 PPG2ABP: translating photoplethysmogram (PPG) signals to arterial blood pressure (ABP) waveforms \emph{Bioengineering} \textbf{9} 692 (doi: 10.3390/bioengineering9110692)

\bibitem[ISO(2018)]{iso8106022018}
International Organization for Standardization 2018 \emph{ISO 81060-2:2018 Non-Invasive Sphygmomanometers---Part 2: Clinical Investigation of Intermittent Automated Measurement Type} 3rd edn (Geneva: ISO) (available at: \url{https://www.iso.org/standard/73339.html})

\bibitem[ISO(2022)]{iso8106032022}
International Organization for Standardization 2022 \emph{ISO 81060-3:2022 Non-Invasive Sphygmomanometers---Part 3: Clinical Investigation of Continuous Automated Measurement Type} 1st edn (Geneva: ISO) (available at: \url{https://www.iso.org/standard/71161.html})

\bibitem[Jeong and Lim(2021)]{jeong2021cnnlstm}
Jeong D U and Lim K M 2021 Combined deep CNN-LSTM network-based multitasking learning architecture for noninvasive continuous blood pressure estimation using difference in ECG-PPG features \emph{Sci. Rep.} \textbf{11} 13539 (doi: 10.1038/s41598-021-92997-0)

\bibitem[Johnson, Pollard, and Mark(2016)]{johnson2016mimicdb}
Johnson A, Pollard T and Mark R 2016 MIMIC-III Clinical Database version 1.4 (PhysioNet) RRID:SCR\_007345 (doi: 10.13026/C2XW26)

\bibitem[Johnson et al.(2016)]{johnson2016mimic}
Johnson A E W, Pollard T J, Shen L, Lehman L H, Feng M, Ghassemi M, Moody B, Szolovits P, Celi L A and Mark R G 2016 MIMIC-III, a freely accessible critical care database \emph{Sci. Data} \textbf{3} 160035 (doi: 10.1038/sdata.2016.35)

\bibitem[Kachuee et al.(2017)]{kachuee2017aami}
Kachuee M, Kiani M M, Mohammadzade H and Shabany M 2017 Cuffless blood pressure estimation algorithms for continuous health-care monitoring \emph{IEEE Trans. Biomed. Eng.} \textbf{64} 859--69 (doi: 10.1109/TBME.2016.2580904)

\bibitem[Kamanditya et al.(2024)]{kamanditya2024convlstm}
Kamanditya B, Fuadah Y N, Mahardika T N Q \emph{et al} 2024 Continuous blood pressure prediction system using Conv-LSTM network on hybrid latent features of photoplethysmogram (PPG) and electrocardiogram (ECG) signals \emph{Sci. Rep.} \textbf{14} 16450 (doi: 10.1038/s41598-024-66514-y)

\bibitem[Kasbekar et al.(2023)]{kasbekar2023optimizing}
Kasbekar R S, Ji S, Clancy E A \emph{et al} 2023 Optimizing the input feature sets and machine learning algorithms for reliable and accurate estimation of continuous, cuffless blood pressure \emph{Sci. Rep.} \textbf{13} 7750 (doi: 10.1038/s41598-023-34677-9)

\bibitem[Kendall et al.(2018)]{kendall2018uncertainty}
Kendall A, Gal Y and Cipolla R 2018 Multi-task learning using uncertainty to weigh losses for scene geometry and semantics \emph{Proc. IEEE/CVF Conf. on Computer Vision and Pattern Recognition} pp 7482--91 (doi: 10.1109/CVPR.2018.00781)

\bibitem[Koenker and Bassett(1978)]{koenker1978quantiles}
Koenker R and Bassett G Jr 1978 Regression quantiles \emph{Econometrica} \textbf{46} 33--50 (doi: 10.2307/1913643)

\bibitem[Lai et al.(2024)]{lai2024dsrunet}
Lai K, Wang X and Cao C 2024 A continuous non-invasive blood pressure prediction method based on deep sparse residual U-Net combined with improved squeeze and excitation skip connections \emph{Sensors} \textbf{24} 2721 (doi: 10.3390/s24092721)

\bibitem[Li et al.(2025)]{li2025fusensering}
Li Z, Lu T, Liu R \emph{et al} 2025 FuSenseRing: an open-source platform for robust cuffless blood pressure monitoring via multimodal sensor fusion and temperature-adaptive attention \emph{Proc. ACM Interact. Mob. Wearable Ubiquitous Technol.} \textbf{9} 192 (doi: 10.1145/3770689)

\bibitem[Liang et al.(2026)]{liang2026cspm}
Liang C \emph{et al} 2026 A conformal piezoelectric microsystem for demographic-adaptive and calibration-free cuffless blood pressure monitoring \emph{Nat. Commun.} \textbf{17} 439 (doi: 10.1038/s41467-025-67118-4)

\bibitem[Lin et al.(2017)]{lin2017focal}
Lin T-Y, Goyal P, Girshick R, He K and Doll\'ar P 2017 Focal loss for dense object detection \emph{Proc. IEEE Int. Conf. on Computer Vision} pp 2980--88 (doi: 10.1109/ICCV.2017.324)

\bibitem[Liu, Wang, et al.(2025)]{liu2024kan}
Liu Z, Wang Y, Vaidya S, Ruehle F, Halverson J, Solja\v{c}i\'{c} M, Hou T Y and Tegmark M 2025 KAN: Kolmogorov--Arnold networks \emph{Proc. Int. Conf. on Learning Representations} (available at: \url{https://proceedings.iclr.cc/paper_files/paper/2025/hash/afaed89642ea100935e39d39a4da602c-Abstract-Conference.html})

\bibitem[Liu, Qiao, et al.(2025)]{liu2025resnetbilstm}
Liu Z, Qiao M, Liu Y, Zhang J and He L 2025 A two-branch ResNet-BiLSTM deep learning framework for extracting multimodal features applied to PPG-based cuffless blood pressure estimation \emph{Sensors} \textbf{25} 3975 (doi: 10.3390/s25133975)

\bibitem[Ma et al.(2023)]{ma2023multiparameter}
Ma G, Zhang J, Liu J, Wang L and Yu Y 2023 A multi-parameter fusion method for cuffless continuous blood pressure estimation based on electrocardiogram and photoplethysmogram \emph{Micromachines} \textbf{14} 804 (doi: 10.3390/mi14040804)

\bibitem[Ma et al.(2026)]{ma2026gaf}
Ma S, Wu Y, Peng C, Zhao Z and Wang H 2026 Continuous blood pressure prediction based on GAF-driven multimodal PPG and ECG signal fusion \emph{Measurement} \textbf{257} 118738 (doi: 10.1016/j.measurement.2025.118738)

\bibitem[McEvoy et al.(2024)]{mcevoy2024esc}
McEvoy J W \emph{et al} 2024 2024 ESC Guidelines for the management of elevated blood pressure and hypertension \emph{Eur. Heart J.} \textbf{45} 3912--4018 (doi: 10.1093/eurheartj/ehae178)

\bibitem[Mehrabadi et al.(2022)]{mehrabadi2022cyclegan}
Mehrabadi M A, Aqajari S A H, Zargari A H A, Dutt N and Rahmani A M 2022 Novel blood pressure waveform reconstruction from photoplethysmography using cycle generative adversarial networks \emph{Proc. 44th Annu. Int. Conf. of the IEEE Engineering in Medicine and Biology Society} pp 1906--09 (doi: 10.1109/EMBC48229.2022.9871962)

\bibitem[Mehta et al.(2024)]{mehta2024challenges}
Mehta S, Kwatra N, Jain M and McDuff D 2024 Examining the challenges of blood pressure estimation via photoplethysmogram \emph{Sci. Rep.} \textbf{14} 18318 (doi: 10.1038/s41598-024-68862-1)

\bibitem[Meidert and Saugel(2018)]{meidert2018noninvasive}
Meidert A S and Saugel B 2018 Techniques for non-invasive monitoring of arterial blood pressure \emph{Front. Med.} \textbf{4} 231 (doi: 10.3389/fmed.2017.00231)

\bibitem[Min et al.(2025)]{min2025wearable}
Min S, An J, Lee J H \emph{et al} 2025 Wearable blood pressure sensors for cardiovascular monitoring and machine learning algorithms for blood pressure estimation \emph{Nat. Rev. Cardiol.} \textbf{22} 629--48 (doi: 10.1038/s41569-025-01127-0)

\bibitem[Moody et al.(2020)]{moody2020mimic3wdb}
Moody B, Moody G, Villarroel M, Clifford G D and Silva I 2020 MIMIC-III Waveform Database version 1.0 (PhysioNet) (doi: 10.13026/c2607m)

\bibitem[Moulaeifard et al.(2025)]{moulaeifard2025generalizable}
Moulaeifard M, Charlton P H and Strodthoff N 2025 Generalizable deep learning for photoplethysmography-based blood pressure estimation---a benchmarking study \emph{Mach. Learn.: Health} \textbf{1} 010501 (doi: 10.1088/3049-477X/ae01a8)

\bibitem[Mukkamala et al.(2015)]{mukkamala2015ptt}
Mukkamala R \emph{et al} 2015 Toward ubiquitous blood pressure monitoring via pulse transit time: theory and practice \emph{IEEE Trans. Biomed. Eng.} \textbf{62} 1879--1901 (doi: 10.1109/TBME.2015.2441951)

\bibitem[Muntner et al.(2019)]{muntner2019bp}
Muntner P, Shimbo D, Carey R M \emph{et al} 2019 Measurement of blood pressure in humans: a scientific statement from the American Heart Association \emph{Hypertension} \textbf{73} e35--e66 (doi: 10.1161/HYP.0000000000000087)

\bibitem[Nawaz et al.(2024)]{nawaz2024transformerabp}
Nawaz M W, Tahir M A, Mehmood A, Rahman M M U, Riaz K and Abbasi Q H 2024 Cuff-less arterial blood pressure waveform synthesis from single-site PPG using Transformer and frequency-domain learning arXiv:2401.05452 (doi: 10.48550/arXiv.2401.05452)

\bibitem[O'Brien et al.(1993)]{obrien1993bhs}
O'Brien E \emph{et al} 1993 Short report: an outline of the revised British Hypertension Society protocol for the evaluation of blood pressure measuring devices \emph{J. Hypertens.} \textbf{11} 677--79 (doi: 10.1097/00004872-199306000-00013)

\bibitem[Pankaj et al.(2026)]{pankaj2025edge}
Pankaj, Maan P, Kumar M, Kumar A and Komaragiri R 2026 Cuffless monitoring of blood pressure using photoplethysmography signal: a comprehensive review of artificial intelligence and edge computing solutions \emph{Arch. Comput. Methods Eng.} \textbf{33} 3837--66 (doi: 10.1007/s11831-025-10415-4)

\bibitem[Pilz et al.(2024)]{pilz2024cuff}
Pilz N \emph{et al} 2024 Cuff-based blood pressure measurement: challenges and solutions \emph{Blood Press.} \textbf{33} 2402368 (doi: 10.1080/08037051.2024.2402368)

\bibitem[Rastegar et al.(2023)]{huang2023cnnsvr}
Rastegar S, Gholam Hosseini H and Lowe A 2023 Hybrid CNN-SVR blood pressure estimation model using ECG and PPG signals \emph{Sensors} \textbf{23} 1259 (doi: 10.3390/s23031259)

\bibitem[Slapni\v{c}ar et al.(2019)]{slapnicar2019ppg}
Slapni\v{c}ar G, Mlakar N and Lu\v{s}trek M 2019 Blood pressure estimation from photoplethysmogram using a spectro-temporal deep neural network \emph{Sensors} \textbf{19} 3420 (doi: 10.3390/s19153420)

\bibitem[Song et al.(2026)]{song2026last}
Song J, Liu Z, Huang Y and Wu X 2026 LAST-CBPM: photoplethysmography-based quasi-continuous blood pressure monitoring algorithm \emph{Biomed. Signal Process. Control} \textbf{120} 109723 (doi: 10.1016/j.bspc.2026.109723)

\bibitem[Stahlschmidt et al.(2022)]{stahlschmidt2022fusion}
Stahlschmidt S R, Ulfenborg B and Synnergren J 2022 Multimodal deep learning for biomedical data fusion: a review \emph{Brief. Bioinform.} \textbf{23} bbab569 (doi: 10.1093/bib/bbab569)

\bibitem[Tang et al.(2024)]{tang2024kd}
Tang H \emph{et al} 2024 Blood pressure estimation based on PPG and ECG signals using knowledge distillation \emph{Cardiovasc. Eng. Technol.} \textbf{15} 39--51 (doi: 10.1007/s13239-023-00695-x)

\bibitem[Tang et al.(2026)]{tang2026fewshot}
Tang L, Lin W-H, Zheng D and Chen F 2026 Enhancing few-shot personalized cuffless blood pressure estimation with self-supervised learning \emph{Physiol. Meas.} \textbf{47} 035012 (doi: 10.1088/1361-6579/ae52a1)

\bibitem[Tanveer and Hasan(2019)]{tanveer2018annlstm}
Tanveer M S and Hasan M K 2019 Cuffless blood pressure estimation from electrocardiogram and photoplethysmogram using waveform based ANN-LSTM network \emph{Biomed. Signal Process. Control} \textbf{51} 382--92 (doi: 10.1016/j.bspc.2019.02.028)

\bibitem[Tian et al.(2025)]{tian2025pctn}
Tian Z, Liu A, Zhu G and Chen X 2025 A paralleled CNN and Transformer network for PPG-based cuff-less blood pressure estimation \emph{Biomed. Signal Process. Control} \textbf{99} 106741 (doi: 10.1016/j.bspc.2024.106741)

\bibitem[Tian et al.(2026)]{tian2026mcaf}
Tian Z, Liu A, Chen J, Wang D and Chen X 2026 PPG-based continuous arterial blood pressure estimation via multi-scale cross attention fusion \emph{Biomed. Signal Process. Control} \textbf{113} 108833 (doi: 10.1016/j.bspc.2025.108833)

\bibitem[Vaswani et al.(2017)]{vaswani2017attention}
Vaswani A \emph{et al} 2017 Attention is all you need \emph{Advances in Neural Information Processing Systems} \textbf{30} 5998--6008

\bibitem[Weber-Boisvert et al.(2023)]{elgendi2023icu}
Weber-Boisvert G, Gosselin B and Sandberg F 2023 Intensive care photoplethysmogram datasets and machine-learning for blood pressure estimation: generalization not guarantied \emph{Front. Physiol.} \textbf{14} 1126957 (doi: 10.3389/fphys.2023.1126957)

\bibitem[Ye et al.(2025)]{difftransformer2024}
Ye T, Dong L, Xia Y, Sun Y, Zhu Y, Huang G and Wei F 2025 Differential Transformer \emph{Proc. Int. Conf. on Learning Representations} (available at: \url{https://proceedings.iclr.cc/paper_files/paper/2025/hash/00b67df24009747e8bbed4c2c6f9c825-Abstract-Conference.html})

\bibitem[Zhao et al.(2023)]{zhao2023emerging}
Zhao L \emph{et al} 2023 Emerging sensing and modeling technologies for wearable and cuffless blood pressure monitoring \emph{npj Digit. Med.} \textbf{6} 93 (doi: 10.1038/s41746-023-00835-6)

\bibitem[Zheng et al.(2025)]{zheng2025utransbpnet}
Zheng Y \emph{et al} 2025 UTransBPNet for cuffless and calibration-free blood pressure estimation under dynamic conditions \emph{Sci. Rep.} \textbf{15} 17654 (doi: 10.1038/s41598-025-02963-3)

\bibitem[Zhou et al.(2021)]{zhou2021global}
Zhou B, Perel P, Mensah G A \emph{et al} 2021 Global epidemiology, health burden and effective interventions for elevated blood pressure and hypertension \emph{Nat. Rev. Cardiol.} \textbf{18} 785--802 (doi: 10.1038/s41569-021-00559-8)

\end{thebibliography}
\end{document}